\documentclass[journal,10pt]{IEEEtran}
\usepackage[utf8]{inputenc}

\usepackage{amsmath}
\usepackage{amssymb}
\usepackage{amsfonts}
\usepackage{cite}

\usepackage[caption=false]{subfig}
\usepackage{booktabs}
\usepackage{bbm}
\usepackage{mathrsfs}

\DeclareMathOperator*{\argmin}{arg\,min}

\newcommand{\Rmnum}[1]{\expandafter\@slowromancap\romannumeral #1@}
\usepackage{graphicx}

\usepackage{epsfig}
\usepackage[numbers,sort&compress]{natbib}
\usepackage{xcolor}
\usepackage{color}
\usepackage{enumerate}
\usepackage{extarrows}
\usepackage[ruled,linesnumbered,lined]{algorithm2e}  
\usepackage{algorithmic}
\usepackage{tensor}
\usepackage[colorlinks,linkcolor=blue,anchorcolor=blue,citecolor=blue,urlcolor=black]{hyperref}
\usepackage{url}
\usepackage{eso-pic}

\begin{document}

\title{Stochastic Quantum Spiking Neural Networks with Quantum Memory and Local Learning }
\author{Jiechen Chen,  \IEEEmembership{Member,~IEEE}, ~Bipin Rajendran,~\IEEEmembership{Senior Member,~IEEE}, and~Osvaldo Simeone,~\IEEEmembership{Fellow,~IEEE}
\thanks{J. Chen is with the Department of Engineering, King’s College London, London, WC2R 2LS, UK (email:jiechen.chen@kcl.ac.uk). B. Rajendran and O. Simeone are with the Institute for Intelligent Networked Systems,
Northeastern University London, One Portsoken Street, London, E1 8PH,
UK (email: \{b.rajendran, o.simeone\}@nulondon.ac.uk).  \\
This work was supported by the European Research Council (ERC) under the European Union’s Horizon Europe Programme (grant agreement No. 101198347), by an Open Fellowship of the EPSRC (EP/W024101/1), and by the EPSRC project (EP/X011852/1).

The authors are grateful to Andrea Gentile (Pasqal) for the useful feedback and advice on hardware implementation.
 }
 \vspace*{-0.6cm}
 }

\maketitle

\AddToShipoutPictureFG*{%
  \AtPageUpperLeft{%
    \raisebox{-12mm}[0pt][0pt]{%
      \hspace{12mm}%
      {\small Published in \textit{IEEE Journal on Selected Areas in Communications}}%
    }%
  }%
}

\IEEEpeerreviewmaketitle
\newtheorem{definition}{\underline{Definition}}[section]
\newtheorem{fact}{Fact}
\newtheorem{assumption}{Assumption}
\newtheorem{theorem}{\underline{Theorem}}[section]
\newtheorem{lemma}{\underline{Lemma}}[section]
\newtheorem{proposition}{\underline{Proposition}}[section]
\newtheorem{corollary}[proposition]{\underline{Corollary}}
\newtheorem{example}{\underline{Example}}[section]
\newtheorem{remark}{\underline{Remark}}[section]
\newcommand{\mv}[1]{\mbox{\boldmath{$ #1 $}}}
\newcommand{\mb}[1]{\mathbb{#1}}
\newcommand{\Myfrac}[2]{\ensuremath{#1\mathord{\left/\right.\kern-\nulldelimiterspace}#2}}
\newcommand\Perms[2]{\tensor[^{#2}]P{_{#1}}}
\newcommand{\note}[1]{[\textcolor{red}{\textit{#1}}]}

\begin{abstract}
Neuromorphic and quantum computing have recently emerged as promising paradigms for advancing artificial intelligence, each offering complementary strengths. Neuromorphic systems built on spiking neurons excel at processing time series data efficiently through sparse, event-driven computation, consuming energy only upon input events. Quantum computing, on the other hand, operates on state spaces that grow exponentially in dimension with the number of qubits --- as a consequence of tensor-product composition --- with quantum states admitting superposition across basis states and entanglement between subsystems. Hybrid approaches combining these paradigms have begun to show potential, but existing quantum spiking models have important limitations. Notably, they implement classical memory mechanisms on single qubits, requiring repeated measurements to estimate firing probabilities, while relying on conventional backpropagation for training. In this paper, we propose a novel stochastic quantum spiking (SQS) neuron model that addresses these challenges. The SQS neuron uses multi-qubit quantum circuits to realize a spiking unit with internal quantum memory, enabling event-driven probabilistic spike generation in a single shot during inference. Furthermore, we study networks of SQS neurons, dubbed SQS neural networks (SQSNN), and demonstrate that they can be trained via a hardware-friendly local learning rule, eliminating the need for global classical backpropagation. The proposed SQSNN model is shown via experiments with both conventional and neuromorphic datasets to improve over previous quantum spiking neural networks, as well as over classical counterparts, when fixing the overall number of trainable parameters, highlighting its potential for event-driven applications such as neuromorphic integrated sensing and communications (N-ISAC).
\end{abstract}


\section{Introduction}\label{sec:intro}
\subsection{Context and Motivation}
Over the past few decades, \emph{quantum computing} and \emph{neuromorphic computing} have emerged as two prominent paradigms for advancing computation beyond the limitations of conventional transistor-based digital architectures \cite{rajendran2023towards, markovic2020quantum}. Each of these technologies brings complementary advantages. \emph{Quantum computing} operates on multi-qubit state spaces whose dimension scales exponentially with the number of qubits due to tensor product composition (see, e.g., \cite{simeone_cqit}). Within this exponentially large space, superposition and entanglement underlie transformations that may not be efficiently simulable on classical computers \cite{bravyi2018quantum, nielsen2010quantum}.

\emph{Neuromorphic computing}, on the other hand, is inspired by the operational principles of biological neural systems. It implements networks of spiking neurons that communicate via discrete, asynchronous events and remain inactive in the absence of input \cite{davies2021advancing}. This event-driven, sparse mode of computation can lead to highly energy-efficient processing, particularly for time-series and streaming data \cite{simeone2026modern}. Neuromorphic computing is highly synergistic with event-driven sensing, an energy-efficient form of sensing that produces information only when sufficiently large changes are detected in the scene. Furthermore, spiking neural networks (SNNs) naturally support online and continual learning, adapting to new information without retraining from scratch \cite{rajendran2019low, mehonic2020memristors, davies2021advancing, skatchkovsky2022bayesian}.

Given these complementary strengths, a compelling question is whether combining quantum and neuromorphic approaches can yield new computational models that inherit the advantages of both domains \cite{markovic2020quantum,brand2024quantum}. An ideal version of this integration would leverage quantum systems to provide exponentially large signal spaces or dynamical richness, while adopting neuromorphic principles to endow the computation with robustness, sparsity, and event-driven efficiency.

Recent works have begun exploring this intersection. For instance, quantum reservoir computing (QRC) leverages the complex dynamics of quantum systems as a reservoir for temporal information processing \cite{dudas2023quantum, gyurik2025quantum}. For example, the work showed that a small network of only 6–7 qubits with disordered quantum evolution can achieve sequence learning performance on par with a classical recurrent neural network of hundreds of neurons \cite{fujii2017harnessing, lau2024modular, murauer2025feedback}.

Another line of work has adapted classical spiking neuron models to quantum hardware. Notably, prior work has  introduced the quantum leaky integrate-and-fire neuron (QLIF) as a unit for the design of quantum spiking neural networks \cite{brand2024quantum}. The QLIF neuron is implemented as a  single-qubit that mimics the leaky integrate-and-fire behavior of classical spiking neurons \cite{davies2021advancing}. By cascading QLIF neurons, this work demonstrated the first \emph{quantum SNN} (QSNN), which was able to perform simple image recognition tasks such as MNIST classification.

\begin{figure*}[t!]
	\centering
	\includegraphics[width=6.3in]{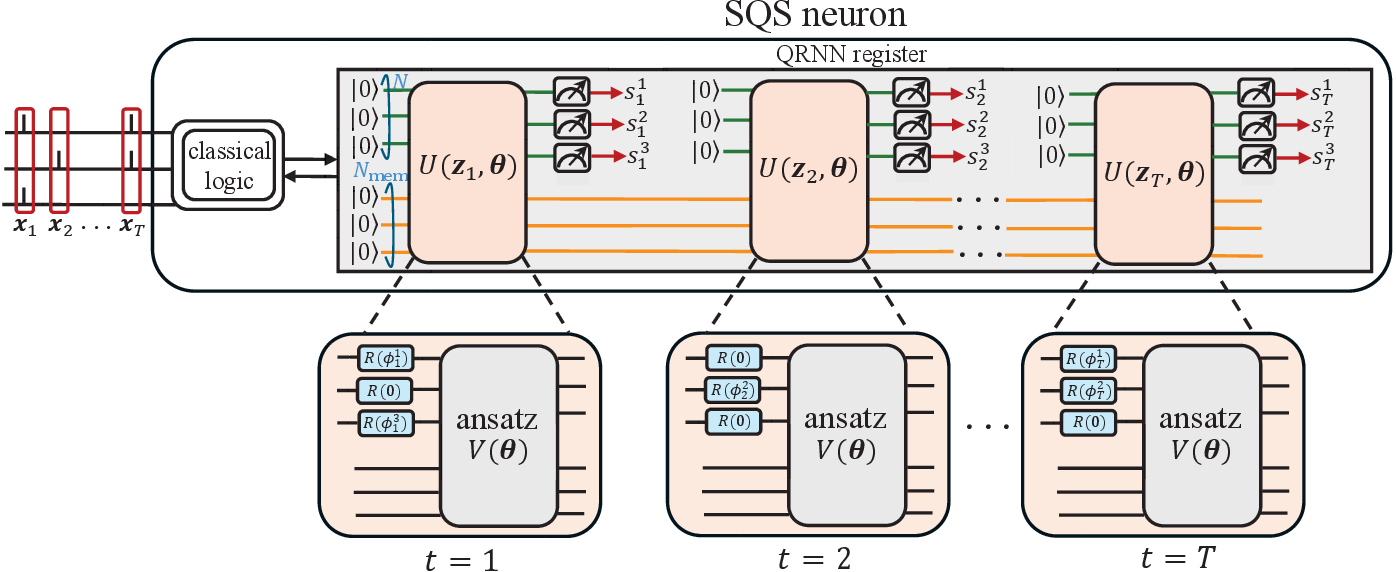}
	\caption{Schematic of an SQS neuron with $N=3$ input-output qubits (green) and $N_{\rm mem}=3$ memory qubits (orange), receiving input from $P$ presynaptic neurons. At each time $t$, incoming presynaptic spikes are weighted into input currents $\mv z_t = [z_t^1,\ldots,z_t^N]$. The currents determine rotation angles $\mv \phi_t$ applied to the input qubits (blue gates). A parameterized quantum circuit (PQC, gray box) with trainable parameters $\mv \theta$ entangles input and memory qubits, updating the neuron’s state. The $N$ input qubits are then measured (meter symbols) to produce output spikes $\mv s_t$ (red arrows), after which those qubits are reset to the ground state $|0\rangle$. The memory qubits are not measured, preserving quantum information (orange shading) that influences subsequent time steps. This internal quantum memory allows the SQS neuron to exhibit complex temporal dynamics, while not requiring  multi-shot measurements.  }
    \vspace{-0.4cm}
	\label{gmodel}
\end{figure*} 

This seminal work proved the viability of spiking neural networks on gate-based quantum computers. However, the QLIF neuron has several limitations from its design:
\begin{itemize}
    \item \emph{Single qubit and classical memory:} First, each QLIF neuron is essentially a classical LIF unit emulated on a single qubit, which restricts it to limited state dynamics characterized by classical state variables.
    \item \emph{Multi-shot measurements:} The QLIF neuron's internal state, analogous to a membrane potential, is encoded as the excitation probability of one qubit, and a threshold rule is used to determine spike emission. In practice, determining whether the neuron should fire requires estimating this excitation probability at each time step via repeated quantum measurements (multiple shots) and comparing against a threshold. This multi-shot measurement scheme adds significant overhead, as the quantum state must be prepared and measured many times to obtain the spiking probability with high confidence. This means the spiking outcome is not generated within a single quantum evolution, but rather via a classical post-processing of measurement statistics.
    \item \emph{Classical training via backpropagation:} Third, the QLIF network training process remains classical, leveraging backpropagation through time on a classical simulation of the network. This reliance on classical, global learning rules makes it challenging to deploy the model directly on quantum hardware for online learning as is the case for classical neuromorphic models \cite{davies2021advancing}.
\end{itemize}

Overall, existing hybrid models like QRC and QLIF validate the concept of quantum neuromorphic systems, but they either treat the quantum device as a black-box dynamical system or implement spiking neurons in a way that closely resembles classical  models. There is a clear need for new designs that fully exploit quantum dynamics while supporting efficient operation and learning on hardware.

\subsection{Main Contributions}
In this work, we address the limitations of the QLIF neuron discussed above by proposing the \emph{Stochastic Quantum Spiking} (SQS) neuron model illustrated in Fig. \ref{gmodel}, along with the corresponding \emph{SQS neural networks} (SQSNNs). The main features of SQS models and SQSNNs are as follows:
\begin{itemize}
    \item \emph{Multi-qubit neuron architecture with quantum memory:} Each SQS neuron is realized using a multi-qubit quantum circuit, allowing the neuron to exploit entangled states and richer internal dynamics. This design increases the expressive capacity of each neuron and ensures that the model truly leverages the unique features of quantum hardware. Specifically, the SQS neuron employs an internal quantum memory mechanism inspired by \emph{quantum recurrent neural networks} (QRNNs) \cite{takaki2021learning, nikoloska2023time}. Accordingly, the neuron’s multi-qubit circuit carries over quantum information from one time instant to the next, preserving relevant memory of the neuron's previous inputs and  activity. Unlike standard QRNNs, however, the proposed SQS neuron uses quantum memory to realize stochastic spike generation from presynaptic spike events, enabling event-driven communication and local learning. Furthermore, SQSNNs support arbitrary connectivity among SQS neurons, making it possible to deploy flexible architectures compatible with state-of-the-art classical SNNs.
    \item \emph{Event-driven, single-shot spike generation:} The SQS neurons produces output spikes \emph{probabilistically} in an event-driven manner based on a \emph{single-shot measurement} of the multi-qubit state. In contrast to QLIF’s multi-shot probability estimation, an SQS neuron emits spikes within one quantum evolution, with the spike event occurring stochastically according to the Born rule. This event-driven operation aligns with neuromorphic principle that no computation should be carried out when no events occurs,  while avoiding costly repeated measurements.
    \item \emph{Hardware-compatible local learning rule:} We develop a novel learning rule for networks of SQS neurons that updates synaptic weights using only locally available information at each SQS neuron, thus avoiding the need for backpropagation on a classical simulator. This bio-inspired learning scheme is designed to run natively on quantum hardware, adjusting connection strengths based on stochastic spike correlations in a manner akin to three-factor learning rules in neuromorphic computing \cite{8891810}. 
\end{itemize}

\begin{figure*}[t!]
	\centering
	\includegraphics[width=6.2in]{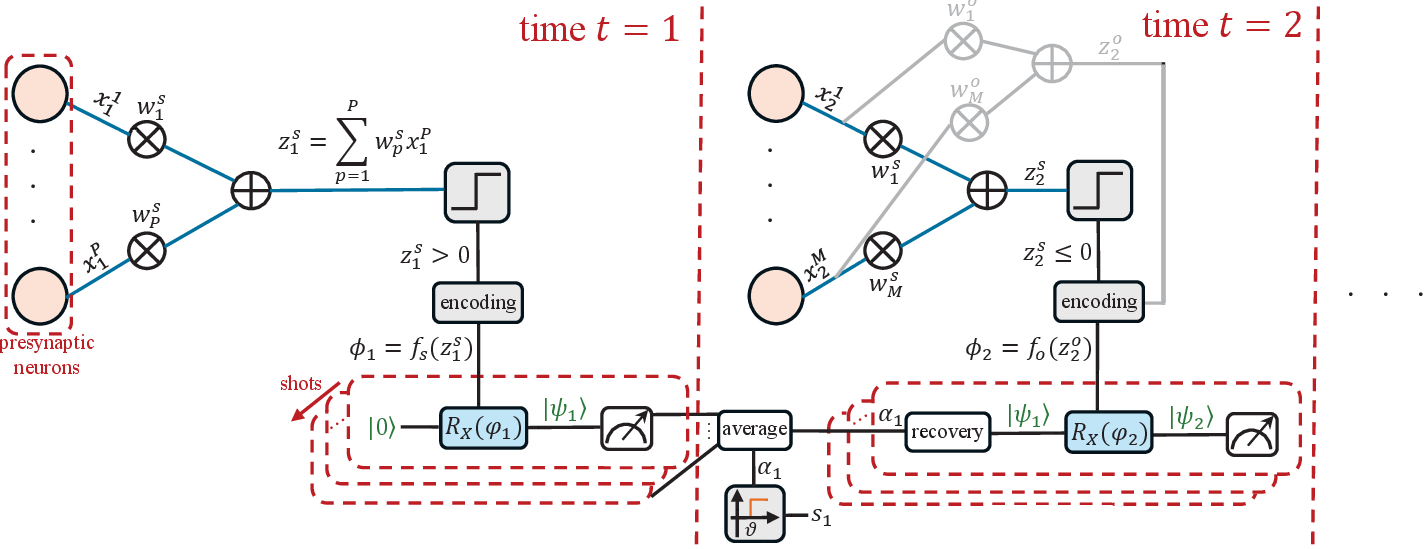}
	\caption{Schematic of the QLIF neuron with $P$ presynaptic inputs \cite{brand2024quantum}. A single qubit (green) serves as the neuron’s internal state, evolving in discrete time steps via a measure-and-prepare scheme. At each time $t$, input spikes from $P$ presynaptic neurons are weighted (blue arrows) and encoded as a rotation $\phi_t$ on the qubit. The qubit is measured multiple times to estimate a spike probability (red arrow), which is compared against a threshold to decide if the neuron fires. This design requires repeated quantum measurements and uses a classical memory (accumulated threshold voltage), mimicking a leaky integrate-and-fire process. }
    \vspace{-0.5cm}
	\label{qlif}
\end{figure*}

Overall, the SQS neuron model harnesses quantum resources, including multiple qubits, entanglement, and inherent measurement randomness, to implement a spiking neuron that operates in an event-driven, probabilistic fashion like a biological neuron. In this regard, SQSNNs can be viewed as examples of dynamic PQCs, which have been recently shown to have significant advantages in terms of expressivity and trainability \cite{deshpande2024dynamic}.

At a practical level, the proposed neural architecture is particularly well suited for  hardware platforms characterized by: (\emph{i}) \emph{low repetition rate}, i.e., high shot-to-shot latency, favoring single-shot measurements \cite{gentile2017noise}; and (\emph{ii}) support for \emph{mid-circuit measurements}  \cite{rudinger2022characterizing},  which are required to implement the QRNN-based  quantum memory and readout mechanisms. Both properties apply to leading candidate platforms for scalable quantum computing, namely trapped ions and neutral atoms \cite{cain2026shor, preskill2025beyond, graham2023midcircuit, yu2025situ}. Furthermore, in practice, the implementation of SQSNNs is subject to the connectivity constraints of the underlying quantum hardware. For example, superconducting quantum processors typically require hardware-aware routing and compilation strategies due to fixed qubit connectivity \cite{cowtan2019qubit}, whereas trapped-ion and neutral-atom platforms offer more flexible qubit interactions and reconfiguration capabilities \cite{tan2024compiling, webber2020efficient}.

Experimental results compare different models by fixing the same number of trainable parameters across all models. This has become a standard benchmarking methodology in quantum machine learning (see, e.g., \cite{xiao2023practical, abbas2021power, fellner2025quantum}), as the number of parameters directly relates to key model properties such as trainability, complexity, and expressivity \cite{abbas2021power}. Our results demonstrate the effectiveness of SQSNNs, and of the local training rule on quantum hardware, as well as its superior accuracy compared to QLIF on a larger-scale dataset trained using backpropagation on a classical simulator.

The rest of the paper is organized as follows. Section \ref{section: qlif} reviews the QLIF neuron proposed by reference \cite{brand2024quantum}. The SQS neuron is proposed in Section \ref{sec:System Model}, while Section \ref{training_back} presents a local training rule of SQS-based neural networks. Experimental setting and results are described in Section \ref{exp}. Finally, Section \ref{con} concludes the paper.

\section{Prior Work: Quantum Leaky Integrate-And-Fire Neuron} \label{section: qlif}
In this section, we review the \emph{quantum leaky integrate-and-fire} (QLIF) neuron proposed by  \cite{brand2024quantum}. As illustrated in Fig.~\ref{qlif}, a QLIF neuron is modeled via an internal single qubit, which evolves over discrete time $t=1,2\ldots$ through a \emph{measure-and-prepare} mechanism characterized by \emph{multi-shot} measurements. The state of the qubit serves as the inner state of the spiking neurons. Furthermore, a rotation angle $\varphi_t$ determines whether the neuron produces a spike or not at time $t$ through a measurement probability evaluated via multiple shots. Consider a QLIF neuron that is fed by $P$ pre-synaptic neurons, at each time $t$, the QLIF neuron receives an input binary signal $\mv x_t=[x_{t}^1, \ldots, x_t^P]^T\in\{0,1\}^P$, with $x_t^p\in\{0,1\}$ representing the binary activation of the $p$-th presynaptic neuron. A signal $x_t^p=1$ indicates the presence of a spike, and $x_t^p=0$ indicates that the $p$-th presynaptic neuron does not spike at time $t$.

The QLIF neuron connects to $P$ presynaptic neurons via two vectors of weights $\mv w^{\rm s}=[w^{\rm s}_1,\ldots, w^{\rm s}_P]^T$ and $\mv w^{\rm o}=[w^{\rm o}_1,\ldots, w^{\rm o}_P]^T$. Specifically, the pre-synaptic signals are aggregated into the input signals
\begin{align}
    z^{\rm s}_t = \sum_{p=1}^P w_p^{\rm s} x_t^p, \label{current1}
\end{align}
and 
\begin{align}
    z^{\rm o}_t = \sum_{p=1}^P w_p^{\rm o} x_t^p. \label{current2}
\end{align}
As in classical neuromorphic computing, the complexity of evaluating the signals \eqref{current1} and \eqref{current2} is proportional to the number of pre-synaptic spikes. Therefore, sparser input signals reduce the complexity of computing \eqref{current1} and \eqref{current2}.

The signals \eqref{current1} and \eqref{current2} are used to determine the state of the internal qubit of the QLIF neuron through a rotation angle $\varphi_t$. In particular, at each time $t$, the internal qubit state $|\psi_t\rangle$ is prepared from the ground state $|0\rangle$ via the Pauli X-rotation gate 
\begin{align}
    R_X(\varphi_t) = \begin{bmatrix}
                    \cos(\varphi_t/2) & -i\sin(\varphi_t/2) \\
                    -i\sin(\varphi_t/2) & \cos(\varphi_t/2)
                \end{bmatrix} \label{xgate}
\end{align}
as $|\psi_t\rangle=R_X(\varphi_t) |0\rangle$, where the angle $\varphi_t$ depends on the input signals \eqref{current1}-\eqref{current2} as discussed below. A standard measurement of the qubit yields output $1$ with probability 
\begin{align}
    \alpha_{t} = |\langle 1|\psi_{t}\rangle|^2 = (\sin(\varphi_t/2))^2. \label{membrane}
\end{align}
This probability is estimated by preparing and measuring state $|\psi_{t}\rangle$ multiple times. The probability \eqref{membrane} represents the counterpart of the membrane potential of classical spiking neurons \cite{10606014,8891810}. In fact, using the estimated probability $\alpha_t$, a spike is generated if the probability $\alpha_t$ exceeds a predetermined threshold $\vartheta\in [0,1]$, i.e., 
\begin{align}
    s_{t}=\left\{
\begin{array}{ll}
0 ~\text{(no spike)},       & \text{if}~ \alpha_t \leq \vartheta, \\
1 ~\text{(spike)},         & \text{if}~ \alpha_t > \vartheta.
\end{array} \right. \label{eq: conventional_spike}
\end{align}
This signal serves as an input to the post-synaptic neurons connected to the given QLIF neuron for time $t+1$.

We now describe the evaluation of the phase $\varphi_t$ used in the rotation \eqref{xgate}. First, at time $t$, the phase $\varphi_t$ is first initialized as  
\begin{align}
    \varphi_t^o=\left\{
\begin{array}{ll}
2 \arcsin (\sqrt{\alpha_{t}}),       & \text{if}~ s_{t-1}=0, \\
0,         & \text{if}~ s_{t-1}=1.
\end{array} \right. \label{recoveryq}
\end{align}
Using \eqref{membrane}, this operation recovers the phase $\varphi_{t-1}$ if no spike is generated at time $t-1$, i.e., if $s_{t-1}=0$. Otherwise, if the QLIF had spiked at the previous time, i.e., $s_{t-1}=1$, the phase is reset to zero, i.e., $\varphi_t^o=0$. Having initialized the phase $\varphi_t^o$, the rotation angle $\varphi_{t}$ is computed as 
\begin{align}
    \varphi_{t} = \varphi_{t}^o + \phi_{t},
\end{align}
where the angle $\phi_{t}$ depends on the signals \eqref{current1}-\eqref{current2} computing from the input spikes. Note that we set $\varphi_0^o=0$, so that the qubit state at time $t=1$ is $|\psi_1\rangle=R_X(\varphi_1) |0\rangle=R_X(\phi_1) |0\rangle$.

When the input current $z^{\rm s}_t$ is positive, the rotation gate applies a positive angle $\phi_t>0$, exciting the membrane potential and increasing the spiking probability \eqref{membrane}, while otherwise a rotation with a negative angle $\phi_t \leq0$ is applied, reducing the spiking probability \eqref{membrane}. Thus a positive input current $z_t^{\rm s}$ causes the QLIF neuron to integrate its inputs, while a negative input current $z_t^{\rm s} \leq 0$  entails the ``leaking" of the membrane potential \eqref{membrane}. Specifically, given the previous spiking probability $\alpha_{t-1}$, the angle $\phi_t$ is determined as
\begin{equation}
\phi_t=\left\{
\begin{array}{ll}
f_s(z_t^{\rm s}),        & \text{if}~ z^{\rm s}_t > 0, \\
f_o(z_t^{\rm o}),       &   \text{if}~ z^{\rm s}_t \leq 0,
\end{array} \right.  \label{encoding_angle}
\end{equation}
with 
\begin{align}
    f_s(z_t^{\rm s})=\pi z^{\rm s}_t \label{fs}
\end{align}
and 
\begin{align}
    f_o(z_t^{\rm o})=-2\arcsin (\sqrt{\alpha_{t-1} \exp{(-\beta |z^{\rm o}_t|/T_1})}), \label{fo}
\end{align}
where $\beta>0$ is a scaling factor and $T_1$ is the parameter that controls the rate of decay towards the ground state $|0\rangle$.  Referring to \cite{brand2024quantum} for details, function $f_s(z_t^{\rm s})$ describes a linear integration of positive current into the rotation \eqref{encoding_angle}, whereas function $f_o(z_t^{\rm o})$ in \eqref{fo} implements leakage by enforcing an exponential decay of the previous excitation probability $\alpha_{t-1}$ with time constant controlled by parameter $T_1$. Reference \cite{brand2024quantum} optimizes the weights $\mv w^{\rm s}$ and $\mv w^{\rm o}$ by using the  arctan surrogate gradient for the spike generation process \eqref{eq: conventional_spike} via backpropagation on a classical computer.

\section{Stochastic Quantum Spiking Neuron Model}\label{sec:System Model}

As illustrated in Fig.~\ref{gmodel}, a SQS neuron can process and produce $N$ spikes, acting as a multiple-input multiple-output unit. As the QLIF neuron reviewed in the previous section, the SQS neuron is composed of a classical logic module that encodes incoming data into quantum rotation gates. However, rather than relying on a classical memory, an SQS neuron encompasses a QRNN module \cite{takaki2021learning, nikoloska2023time} that plays a role of \emph{quantum memory} through an event-driven parameterized quantum circuit (PQC). Furthermore, unlike the QLIF neuron, the SQS neuron only requires single-shot measurements, thus foregoing the need for multiple measurement-and-prepare steps. This enables the construction of quantum SNN models in which inference requires single-shot measurements at all neurons, with the neurons communicating via spiking signals.

\subsection{Architecture of the SQS Neuron}
 As illustrated in Fig.~\ref{gmodel}, the SQS neuron encompasses two groups of qubits: $N$ input-output qubits and $N_{\rm mem}$ memory qubits for a total of $N_{\rm tot}=N+N_{\rm mem}$ qubits. The input-output qubits are reset to the ground state $|0\rangle^{\otimes N}$ at the beginning of each time step $t$, processing the incoming signals from the pre-synaptic neurons and producing output spikes via measurement. The memory qubits evolve over time jointly with the input-output qubits to serve as memory for the SQS neuron. Unlike LIF or QLIF spiking neurons, which maintain a scalar membrane potential governed by first-order integrate-and-leak dynamics, the proposed SQS neuron efficiently evolves a state vector on a $2^{N_{\mathrm{tot}}}$-dimensional Hilbert space through a trainable unitary transformation, potentially enabling a richer class of temporal state transitions. In this sense, the SQS neuron is aligned with recent advances in the design of classical spiking neurons that support complex dynamics via the integration with state-space models (SSMs) \cite{karilanova2025state}.

As shown in Fig.~\ref{gmodel}, at each discrete time $t=1,\ldots, T$, a PQC defined by a parameterized unitary transformation $U(\mv z_{t}, \mv \theta)$ operates on the register of $N+N_{\rm mem}$ qubits, where $\mv \theta$ are parameters of the PQC to be optimized.  Assuming that the neuron is connected to $P$ pre-synaptic neurons, at each time $t$, the neuron processes $NP$ incoming spiking signals $\mv x_t=[\mv x_{t,1}, \ldots, \mv x_{t,P}]^T$ with $\mv x_{t,p}=[x_{t,p}^1, \ldots, x_{t,p}^N]^T \in\{0,1\}^N$ representing the $N$ spiking signals produced by the $p$-th pre-synaptic neuron. The SQS neuron itself produces $N$ spiking signals $\mv s_{t}=[s_{t}^1,\ldots,s_{t}^N]^T\in\{0,1\}^{N}$ by measuring the $N$ input-output qubits through the following steps:
\begin{itemize}
    \item \emph{Quantum embedding:} Convert the $NP \times 1$ input signals $\mv x_t$ into $N \times 1$ input currents $\mv z_{t}=[z_{t}^1, \ldots, z_{t}^N]$, and embed the input currents $\mv z_{t}$ into the state of the input-output qubits via rotation gates;
    \item \emph{Quantum memory retrieval:} Following the QRNN model, evolve jointly input-output qubits and memory qubits;
    \item \emph{Single-shot measurements:} Measure the $N$ input-output qubits to produce the $N \times 1$ spiking signals $\mv s_t$; 
\end{itemize}
Importantly, while the spiking output of the QLIF neuron is determined on the basis of multi-shot measurements via the thresholding function \eqref{eq: conventional_spike}, the SQS neuron is run only once, producing a random $N$-dimensional spiking output on the basis of a single-shot measurement.

These steps are elaborated as in the rest of this section.

\begin{figure}[htp]
	\centering
	\includegraphics[width=2.8in]{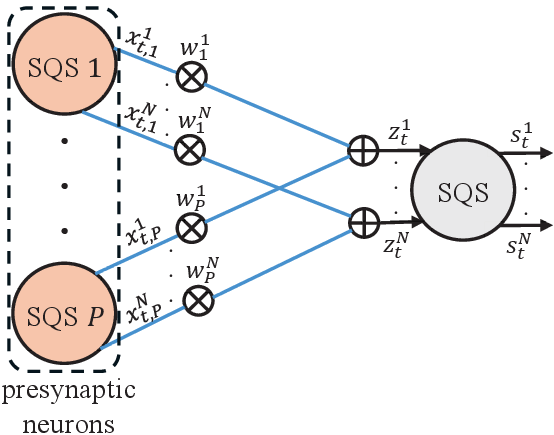}
	\caption{Illustration of the connection between a set of $P$ presynaptic neurons and the SQS neuron of interest. The $n$-th input current $z_{t}^n$ of the SQS neuron is computed as the weighted sum of the corresponding spiking signal $\{x_{t,p}^n\}_{p=1}^P$ from the $P$ presynaptic neurons as in equation \eqref{current}. }
	\label{network}
\end{figure} 

\vspace{-0.3cm}
\subsection{Quantum Embedding}
As illustrated in Fig.~\ref{network}, denote the weights between the $n$-th output of all pre-synaptic neurons and the neuron under study as $\mv w^n=[w_{1}^n, \ldots, w_{P}^n]$, the spikes $\mv x_t=[\mv x_{t,1}, \ldots, \mv x_{t,P}]$ from the $P$ pre-synaptic neurons are weighed to produce the $N$ input currents $\mv z_{t}=[z_{t}^1,\ldots, z_{t}^N]^T$. Writing the $P$ spiking signals from all pre-synaptic neurons' $n$-th output  as $\mv x_{t}^n=[x_{t,1}^n, \ldots, x_{t,P}^n]^T$, the $n$-th element of the current, $z_t^n$, is evaluated as  
\begin{align}
    z_{t}^n = \mv w^n \mv x_{t}^n. \label{current}
\end{align}
As for \eqref{current1} and \eqref{current2}, the complexity of evaluating \eqref{current} is proportional to the number of spikes generated by the pre-synaptic neurons.

Furthermore, the input currents $\mv z_{t}$ are then embedded into rotation angles $\mv \phi_t=[\phi_t^1, \ldots, \phi_t^N]^T$, with the $n$-th angle given by
\begin{align}
    \phi_{t}^n=\left\{
\begin{array}{ll}
f_s(z_{t}^n),      & \text{if}~ z_{t}^n >0, \\
0,         & \text{if}~ z_{t}^n \leq 0,
\end{array} \right. \label{recovery}
\end{align}
with $f_s(z_{t}^n)=\pi z_{t}^n$ for all $n=1, \ldots, N$. Each angle $\phi_t^n$ determines the rotation $R_X(\phi_{t}^n)$ applied to the $n$-th input-output qubit. The synaptic weights $\mv w^n$ are real-valued learnable parameters. Unlike for the QLIF neuron (see \eqref{recoveryq}), the angle $\phi_{t}^n$ of the rotations $R_X(\phi_{t}^n)$ does not implement any memory mechanisms,  since memory effects are accounted for by the memory qubit, as discussed below.

By \eqref{recovery}, when the current $z_{t}^n \leq 0$, the corresponding qubit does not apply any rotation, reducing the number of internal quantum operations required. Therefore, sparsity in the spiking signals $\mv x_t^n$ reduces the complexity of both the classical logic evaluating \eqref{recovery} and the quantum operations applied to the neuron's qubit.

\subsection{Quantum Memory Retrieval}
As discussed in the previous subsection, at each time $t$, the input-output qubits are initialized to the ground state $|0\rangle^{\otimes N}$ and are applied the rotation gates $\{R_X(\phi_{t}^n)\}_{n=1}^N$ to embed the input currents $\mv z_{t}$ via \eqref{current} and \eqref{recovery}. These input-output qubits, along with the memory qubits in the second group, are processed by a PQC defined by a unitary matrix $V(\mv \theta)$. Accordingly, the overall unitary transformation $U(\mv z_{t}, \mv \theta)$ applied by the SQS neuron to the initial state of the $(N+N_{\rm mem})$-qubit register can be expressed as
\begin{align}
    U(\mv z_{t}, \mv \theta)= V(\mv \theta) \cdot \bigg(\bigotimes_{n=1}^N R_X(\phi_{t}^n) \otimes I \bigg), \label{chain0}
\end{align}
where $I$ is a $2^{N_{\rm mem}} \times 2^{N_{\rm mem}}$ identity matrix. 

After joint transformation \eqref{chain0}, the input-output qubits are measured to generate the output spiking signal $\mv s_{t}\in\{0,1\}^N$. In contrast, the memory qubits are not measured, retaining information across time steps.

\subsection{Single-Shot Measurements}
At the initial time $t=1$, all the $N_{\rm tot}$ qubits are all initialized to the ground state $|0\rangle$. Denote as $\rho_t^{\rm IM}$ the density matrix of the $N+N_{\rm mem}$ qubits at the beginning of time interval $t$ and as 
\begin{align}
    \rho^{\rm I}_t=\text{Tr}_{\rm M}(\rho^{\rm IM}_t) \label{input_state}
\end{align}
the corresponding reduced density matrix for the input-output qubits.

By Born's rule, the probability of producing the output signals $\mv s_t \in\{0,1\}^N$ is given by
\begin{align}
    p(\mv s_{t}|\rho_t^{\rm I})= \langle \mv s_{t}| \rho_t^{\rm I} |\mv s_{t} \rangle, \label{prob_spike}
\end{align}
where $|\mv s_{t} \rangle= |s_{t}^1,\ldots, s_t^N \rangle$. 

Furthermore, as the input-output qubits collapse onto the computational state $|\mv s_{t}\rangle$, the post-measurement state of the memory qubits is described by the density matrix
\begin{align}
    \sigma^{\rm mem}_t= \frac{(\langle \mv s_{t}| \otimes I) \rho^{\rm IM}_t(|\mv s_{t} \rangle \otimes I)}{\langle \mv s_{t}| \rho_t^{\rm I} |\mv s_{t} \rangle}, \label{memory_state}
\end{align}
where $I$ is a $2^{N_{\rm mem}} \times 2^{N_{\rm mem}}$ identity matrix.
Accordingly, upon the application of the given unitary matrix \eqref{chain0},  the density matrix of all the qubits at the following time step $t+1$ is  
\begin{align}
    \rho_{t+1}^{\rm IM}=U(\mv z_{t}, \mv \theta)(|0\rangle \langle 0|^{\otimes N} \otimes \sigma^{\rm mem}_t)U(\mv z_{t}, \mv \theta)^\dagger. \label{compose}
\end{align}
Through the relation \eqref{compose}, as anticipated, the state $\sigma^{\rm mem}_t$ of the second group of qubits serves as memory.

\section{Local Training of SQS-based Neural Networks} \label{training_back}
Similar to conventional neural models and to the QLIF neuron, the SQS neuron proposed in the previous section can be structured into a neural network that can be trained to perform specific tasks.  In this section, we first describe a general network architecture based on SQS neurons, and then introduce a training rule designed to run effectively on quantum hardware.  As in neuromorphic systems \cite{davies2021advancing}, the training rule is \emph{local}, in the sense that updates to the parameters can be evaluated at each neuron based on information available locally, along with a scalar global feedback signal \cite{8891810, frenkel2023bottom}. Technically, this result is obtained by leveraging the probabilistic, autoregressive, structure of the model induced by a network of SQS neurons. The local training rule builds on standard zero-th order perturbation-based gradient estimates \cite{simeone2022introduction} within each neuron, while enforcing cooperation across all neurons via global feedback signals.

\subsection{SQS-based Neural Network Model} \label{SQS_model}
\begin{figure}[htp]
	\centering
	\includegraphics[width=2.1in]{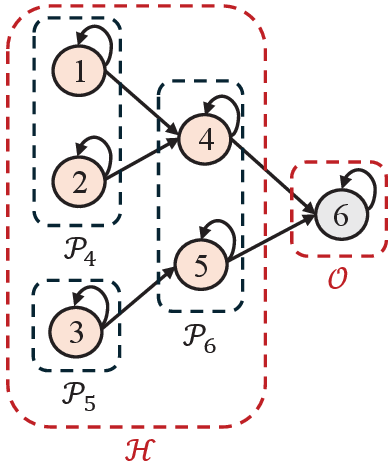}
	\caption{Illustration of an SQS-based neural network (SQSNN) with six neurons ($\mathcal{S}=\{1,\dots,6\}$). Edges $(i \to j)$ indicate neuron $i$ feeds into neuron $j$. Neurons $\{1,2,3,4,5\}$ are hidden, and neuron $6$ is the only output neuron ($\mathcal{O}=\{6\}$). Output neurons have no outgoing edges. Each node is an SQS neuron as in Fig.~\ref{gmodel}, with a self-loop denoting dependence on its own past spikes. The network’s joint spike probability factorizes along these connections: e.g., neuron $6$’s spiking at time $t$ depends only on the past spikes of neurons 4, 5 (its parents, denoted as $\mathcal{P}_6$) and on its own past spikes. }
	\label{graph}
\end{figure}

As shown in Fig.~\ref{graph}, an \emph{SQS-based neural network} (SQSNN) consists of a directed graph $(\mathcal{S}, \mathcal{E})$ encompassing a set $\mathcal{S}$ of SQS neurons. The neurons are connected via directed edges described by a set $\mathcal{E} \subseteq \{(i,j) ~\text{with}~ i,j\in\mathcal{S} ~\text{and}~ i \ne j\}$ of ordered pairs $(i,j)$. Accordingly, each neuron $i$ within the SQSNN receives inputs from a set $\mathcal{P}_i$ of pre-synaptic neurons, with $\mathcal{P}_i=\{j\in\mathcal{S}: j \ne i ~\text{and}~ (j,i)\in\mathcal{E}\}$. The number of pre-synaptic neurons for neuron $i$ is thus $P_i=|\mathcal{P}_i|$. 

Following the literature on neuromorphic systems (see, e.g., \cite{8891810,jang2021vowel}),  we partition the SQS neurons in the SQSNN into two disjoint subsets: the subset $\mathcal{O}$ of output neurons and the subset $\mathcal{H}$ of hidden neurons, with $\mathcal{S}=\mathcal{O} \cup \mathcal{H}$. The spiking signals of the output neurons provide the outputs of the SQSNN. In contrast, the spiking signals of the hidden neurons serve as internal representations guiding the output neurons towards producing informative signals.

Furthermore, as illustrated by the example in Fig.~\ref{graph}, the output neurons do not have any outgoing edges, representing the last layer of the network. Mathematically, we have the condition
\begin{align}
    \mathcal{P}_i \cap \mathcal{O}=\emptyset, \label{no_edge}
\end{align}
or equivalently $\mathcal{P}_i \cap \mathcal{H} = \mathcal{P}_i$, for all neurons $i\in \mathcal{S}$.

Being based on an arbitrary directed graph connecting hidden and output neurons, the architecture of the proposed SQSNN is aligned and compatible with existing SNN architectures \cite{8891810}. Accordingly, SQSNN models can be obtained by replacing classical LIF neurons with SQS neurons, while preserving the original network topology, thus suggesting a natural design strategy for SQSNN architectures.

As a final note, we observe that when the PQCs of all neurons are fixed and not trained, the SQSNN architecture introduced here can be interpreted as a form of quantum reservoir computing (QRC) with mid-circuit measurements \cite{sannia2025exponential,murauer2025feedback,lau2024modular}. In it, the neurons serve as module of the reservoir with classical communication between modules, defined via mid-circuit measurements.

\subsection{Probabilistic Autoregressive Model}
As described in Section \ref{SQS_model}, at each time $t$, each (hidden or output) neuron $i$, receives the spiking signals $\mv s_{\mathcal{P}_i, t}\in\{0,1\}^{NP_i}$ from its pre-synaptic neurons in set $\mathcal{P}_i$, and produces a spiking vector-valued signal $\mv s_{i,t}\in\{0,1\}^N$. Denote the past spiking output of any neuron $i\in\mathcal{S}$ up to time $t-1$ as $\mv s_i^{t-1}=[\mv s_{i,1}, \ldots, \mv s_{i, t-1}]$, and the collection of all past spiking outputs of any subset $\mathcal{S}^{\prime} \subseteq \mathcal{S}$ of neurons as $\mv s_{\mathcal{S}^{\prime}}^{t-1}=\{\mv s_i^{t-1}\}_{i\in\mathcal{S}^{\prime}}$. Following the SQS model introduced in Section \ref{sec:System Model}, the signal $\mv s_{i,t}$ produced by neuron $i$ at any time $t$ depends on the past spiking outputs $\mv s_{\mathcal{P}_i\cup \{i\}}^{t-1}$ from itself and from the pre-synaptic neurons $\mathcal{P}_i$. 

Specifically, the dependence of the spiking history $\mv s_{\mathcal{P}_i\cup \{i\}}^{t-1}$ is reflected by the expression \eqref{prob_spike} of the spiking probability as a function of the neuron's state $\rho^{\rm I}_{i,t}$ of  the input-output qubits of neuron $i$. In fact, by the dynamic update \eqref{compose}, the state $\rho_{i,t}^{\rm I}$ of the input-output qubits of neuron $i$ evolves as a function of the sequence $\mv s_{\mathcal{P}_i\cup \{i\}}^{t-1}$ through the currents \eqref{current}. Accordingly, the spiking probability of neuron $i$ at time $t$, when conditioned on the sequence $\mv s_{\mathcal{P}_i\cup \{i\}}^{t-1}$, can be written using \eqref{prob_spike} as
\begin{align}
    p_{\scalebox{0.7}{\mv \Theta}_i}(\mv s_{i,t}|\mv s_{\mathcal{P}_i\cup \{i\}}^{t-1})= \langle \mv s_{i,t}| \rho_{i,t}^{\rm I} |\mv s_{i,t} \rangle. \label{prob_spike2}
\end{align}
The notation $p_{\scalebox{0.7}{\mv \Theta}_i}(\mv s_{i,t}|\mv s_{\mathcal{P}_i\cup \{i\}}^{t-1})$ emphasizes the dependence of the probability \eqref{prob_spike2} on the local parameters $\mv \Theta_i=\{\mv w_i, \mv \theta_i\}$ of neuron $i$. These include the synaptic weights $\mv w_i=\{\{w_{i,p}^n\}_{p=1}^{P_i}\}_{n=1}^N$ between the $N$ input-output qubits of the $P_i$ pre-synaptic neurons and the corresponding $N$ qubits of neuron $i$, as well as the parameters $\mv \theta_i$ of the PQC in neuron $i$.  Indeed, the state $\rho^{\rm I}_{i,t}$ is obtained via \eqref{input_state} and \eqref{compose}, where the unitary transformation $U(\mv z_{i,t}, \mv \theta_i)$ in \eqref{chain0} depends on the variational parameters $\mv \theta_i$. 

Denote as $\mv \Theta =\{\mv \Theta_i\}_{i\in\mathcal{S}}$ the collection of all model parameters. By applying the chain rule, the joint spiking probability $\mv s^T=[\mv s_1^T, \ldots, \mv s_{|\mathcal{S}|}^T]$ of all neurons in the SQSNN up to time $T$ can be factorized over time index $t$ and over neuron $i$ as
\begin{align}
    p_{\scalebox{0.7}{\mv \Theta}}(\mv s^T) &~= \prod_{t=1}^T p_{\scalebox{0.7}{\mv \Theta}}(\mv s_{t}|\mv s^{t-1}) \notag \\
    &~= \prod_{t=1}^T \prod_{i\in \mathcal{S}} p_{\scalebox{0.7}{\mv \Theta}_i}(\mv s_{i,t}|\mv s_{\mathcal{P}_i\cup \{i\}}^{t-1}). \label{point_prob}
\end{align}
This corresponds to a probabilistic autoregressive sequence model akin to probabilistic spiking models studied in, e.g., \cite{8891810,jang2021vowel}.

Denote the spiking signals emitted by the output neurons as $\mv o_t=\{\mv o_{i,t}\}_{i\in\mathcal{O}}$, with $\mv o_{i,t}=[o_{i,t}^1, \ldots, o_{i,t}^N]$, while the spiking signals emitted by the hidden neurons are written as $\mv h_t=\{\mv h_{i,t}\}_{i\in\mathcal{H}}$, with $\mv h_{i,t}=[h_{i,t}^1, \ldots, h_{i,t}^N]$. As mentioned, the signals $\mv o_t$ correspond to the model output (see Fig.~\ref{graph}).  We also denote as $\mv \Theta^O=\{\mv \Theta_i\}_{i\in\mathcal{O}}$ and $\mv \Theta^H=\{\mv \Theta_i\}_{i\in\mathcal{H}}$ the sets of local parameters for the observed and hidden neurons, respectively. With these definitions, given that, by \eqref{no_edge}, the output neurons have no outgoing edges, the probability \eqref{point_prob} can be expressed as 
\begin{align}
    p_{\scalebox{0.7}{\mv \Theta}}(\mv s^T)= \prod_{t=1}^T p_{\scalebox{0.7}{\mv \Theta}^H}(\mv h_t|\mv h^{t-1}) p_{\scalebox{0.7}{\mv \Theta}^O}(\mv o_t|\mv o^{t-1}, \mv h^{t-1}), \label{all_prob}
\end{align}
where the spiking probability $p_{\scalebox{0.7}{\mv \Theta}^H}(\mv h_t|\mv h^{t-1})$ of the hidden neurons does not depend on the spiking probability of the output neurons due to the assumption \eqref{no_edge}.

\subsection{Maximum Likelihood Learning} \label{mll}
To train an SQSNN, we assume the availability of a dataset $\mathcal{D}=\{\mv o^T\}$ of sequences $\mv o^T=[\mv o_1, \ldots, \mv o_T]$, where each sequence $\mv o^T$ corresponds to an example of desired behavior for the output neurons. Note that the sequence $\mv o^T$ in the training dataset is used to supervise the spiking activity of the output neurons $i\in\mathcal{O}$, while the spiking behavior of the hidden neurons is left unspecified by the data.  In practice, the data typically also specifies the (classical) inputs to a number of neurons \cite{8891810, jang2021vowel}. Examples can be found in \cite{8891810, jang2021vowel} and in Section \ref{exp}.  The inclusion of inputs is not made explicit in set $\mathcal{D}$ to simplify the notation. 

By the standard \emph{maximum likelihood} (ML) criterion, we aim at minimizing the empirical average log-loss on data set $\mathcal{D}$, yielding the optimized model parameters 
\begin{align}
    \mv \Theta^* = \argmin_{\mv \Theta} \bigg\{\mathcal{L}_{\mathcal{D}}(\mv \Theta)=\sum_{\scalebox{0.8}{\mv o}^T\in\mathcal{D}} \mathcal{L}_{\scalebox{0.8}{\mv o}^T}(\mv \Theta)\bigg\}, \label{loss}
\end{align}
with the per-data-point log-loss
\begin{align}
    \mathcal{L}_{\scalebox{0.8}{\mv o}^T}(\mv \Theta) &~= -\log p_{\scalebox{0.7}{\mv \Theta}}(\mv o^T) \notag\\
    &~= -\log \sum_{\scalebox{0.7}{\mv h}^T}p_{\scalebox{0.7}{\mv \Theta}}(\mv s^T), \label{problem}
\end{align}
where $\sum_{\scalebox{0.7}{\mv h}^T}$ denotes the sum over all possible $2^{|\mathcal{H}|}$ values $\mv h^T$ taken by the signals produced by hidden neurons. By \eqref{problem}, evaluating the log-loss $\mathcal{L}_{\scalebox{0.8}{\mv o}^T}(\mv \Theta)$ requires marginalizing over the spiking signals of the hidden neurons.

A direct application of standard gradient-based methods in quantum machine learning to problem \eqref{loss} is not feasible. In fact, this would require estimating the probabilities $p_{\scalebox{0.7}{\mv \Theta}}(\mv s^T)$ in \eqref{problem} for all possible realization $\mv h^T$ of the hidden neurons. To address this problem, we proceed in two steps. First, we introduce a bound on the log-loss \eqref{problem} that enables Monte Carlo estimation, removing the need for the exponential number of network runs required to evaluate the sum in \eqref{problem}. Second, using the bound, we derive a local learning rule based on local perturbation-based gradient estimates and a global scalar feedback signal.

\subsection{Estimating a Likelihood Bound}
To derive the proposed training method, we start by introducing an upper bound on the log-likelihood. As in \eqref{problem}, marginalizing the joint distribution  \eqref{all_prob}, the marginal likelihood over hidden trajectories can be written as
\begin{align}
    p_{\scalebox{0.7}{\mv \Theta}}(\mv o^T)&~=\sum_{\scalebox{0.7}{\mv h}^T}\prod_{t=1}^T p_{\scalebox{0.7}{\mv \Theta}^H}(\mv h_t|\mv h^{t-1}) p_{\scalebox{0.7}{\mv \Theta}^O}(\mv o_t|\mv o^{t-1}, \mv h^{t-1}) \notag\\ &~=\prod_{t=1}^T\mathbb{E}_{p_{\scalebox{0.7}{\mv \Theta}^H}(\scalebox{0.8}{\mv h}_t|\scalebox{0.8}{\mv h}^{t-1})}[p_{\scalebox{0.7}{\mv \Theta}^O}(\mv o_t|\mv o^{t-1}, \mv h^{t-1})].
\end{align}
As a result, the negative log-likelihood can be expressed as
\begin{align}
    \mathcal{L}_{\scalebox{0.8}{\mv o}^T}(\mv \Theta) &~ = -\sum_{t=1}^T \log \mathbb{E}_{p_{\scalebox{0.7}{\mv \Theta}^H}(\scalebox{0.8}{\mv h}_t|\scalebox{0.8}{\mv h}^{t-1})}[p_{\scalebox{0.7}{\mv \Theta}^O}(\mv o_t|\mv o^{t-1}, \mv h^{t-1})] \notag \\
    &~ \leq -\sum_{t=1}^T \mathbb{E}_{p_{\scalebox{0.7}{\mv \Theta}^H}(\scalebox{0.8}{\mv h}_t|\scalebox{0.8}{\mv h}^{t-1})}[\log p_{\scalebox{0.7}{\mv \Theta}^O}(\mv o_t|\mv o^{t-1}, \mv h^{t-1})]  \notag \\
    &~ = L_{\scalebox{0.8}{\mv o}^T}(\mv \Theta), \label{lower_bound}
\end{align}
where the second line follows from Jensen's inequality \cite{8891810}. Using the factorization of the conditional spike distribution in \eqref{point_prob}, the upper bound $L_{\scalebox{0.8}{\mv o}^T}(\mv \Theta)$ in \eqref{lower_bound} can be rewritten as
\begin{align}
    L_{\scalebox{0.8}{\mv o}^T}(\mv \Theta) &~= -\sum_{t=1}^T \sum_{i\in\mathcal{O}} \mathbb{E}_{p_{\scalebox{0.7}{\mv \Theta}^H}(\scalebox{0.8}{\mv h}_t|\scalebox{0.8}{\mv h}^{t-1})}[\log p_{\scalebox{0.7}{\mv \Theta}_i^O}(\mv o_{i,t}|\mv o_{\mathcal{P}_i\cup \{i\}}^{t-1})] \notag \\
    &~ = -\sum_{t=1}^T \sum_{i\in\mathcal{O}} \mathbb{E}_{p_{\scalebox{0.7}{\mv \Theta}^H}(\scalebox{0.8}{\mv h}_t|\scalebox{0.8}{\mv h}^{t-1})}\big[\log \big( \text{Tr}(\rho_{i,t}^{\rm I}O_{i,t}) \big) \big], \label{upper}
\end{align}where $\text{Tr}(\cdot)$ is the trace operation, and the observable $O_{i,t}$ is defined as
\begin{align}
    O_{i,t}= |\mv o_{i,t}\rangle \langle \mv o_{i,t}|.
\end{align}

The key advantages of the upper bound $L_{\scalebox{0.8}{\mv o}^T}(\mv \Theta)$ in \eqref{upper} is that it can be approximated via a Monte-Carlo estimate obtained by running forward passes on the model, with each forward pass producing random samples $\mv h_t \sim p_{\scalebox{0.7}{\mv \Theta}^H}(\scalebox{0.8}{\mv h}_t|\scalebox{0.8}{\mv h}^{t-1})$ for the spiking signals of the hidden neurons. 

We are interested in minimizing the upper bound on the training loss $\mathcal{L}_{\mathcal{D}}(\mv \Theta)$ in \eqref{loss} given by $L_{\mathcal{D}}(\mv \Theta)=\sum_{\scalebox{0.8}{\mv o}^T\in\mathcal{D}} L_{\scalebox{0.8}{\mv o}^T}(\mv \Theta)$, i.e.,
\begin{align}
    \mv \Theta^* = \argmin_{\mv \Theta} \bigg\{L_{\mathcal{D}}(\mv \Theta)=\sum_{\scalebox{0.8}{\mv o}^T\in\mathcal{D}} L_{\scalebox{0.8}{\mv o}^T}(\mv \Theta)\bigg\}. \label{bound_problem}
\end{align}

\subsection{Local Zero-th Order Learning}
Given the objective \eqref{bound_problem}, a conventional-based scheme would estimate the gradient $\nabla_{\scalebox{0.7}{\mv \Theta}} L_{\scalebox{0.8}{\mv o}^T}(\mv \Theta)$ via local perturbations of \emph{all} parameters \cite{spall1992multivariate}. However, this would require a number of $\emph{global}$ forward passes through the entire network that grows \emph{linearly} with the number $|\mv \Theta|$ of parameters in $\mv \Theta$ \cite{spall1992multivariate}. This becomes quickly impractical as the number of neurons grows.

To address the computational and scalability challenges of conventional zeroth-order training, in this subsection, we propose a local learning rule in which each neuron updates its parameters using a single set of $M$ \emph{global} forward trajectories, followed by \emph{local} forward passes at each neuron for gradient estimation. Through this method, the number of global forward executions is reduced from a quantity proportional to the number of parameters  $|\mv \Theta|$ to the constant $M$. This relocation from global passes to local computations facilitates scalable training while preserving correct gradient estimation. 

\begin{figure}[htp]
	\centering
\includegraphics[width=3.4in]{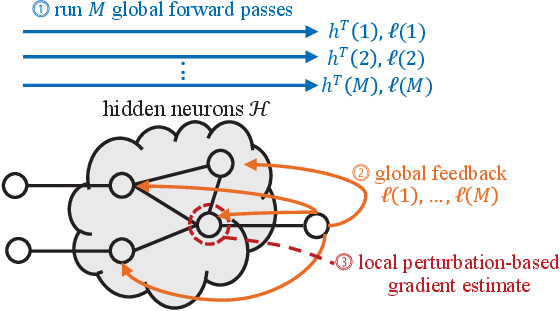}
	\caption{Illustration of the proposed local zero-th order learning rule. At each iteration, the proposed local zeroth-order learning rule implements: \textcircled{1} $M$ global forward passes, where $M$  is  a constant $M$ independent of the number of parameters, \textcircled{2} the evaluation of $M$ global feedback signals at the output SQS neurons, and \textcircled{3} local perturbation-based, zeroth-order updates within each neuron.}
	\label{local}
\end{figure}

As illustrated in Fig.~\ref{local}, given the current iterate $\mv \Theta$ and a data point $\mv o^T$, we begin by running the SQSNN with parameters $\mv \Theta$ for $M$ forward passes to collect $M$ samples $\mv h^T(1), \ldots, \mv h^T(M)$ of the hidden spiking signals. After this initial phase, each neuron updates its local parameters using zeroth-order perturbation methods that require only \emph{local} forward passes within each neuron. Finally, the model parameters are updated at each neuron based on the results of the local forward passes and the \emph{global feedback signals} $\ell(1), \ldots, \ell(M)$.

Accordingly, unlike conventional zero-th order learning methods, which require a separate set of forward passes for each  parameter (see e.g., \cite{schuld2021machine, simeone2022introduction}), our method performs just one set of $M$ forward runs, after which local updates are computed independently for each neuron.

To derive the local gradient updates within each neuron, we consider separately hidden and output neurons.
\subsubsection{Output neurons}
Using \eqref{upper}, the gradient with respect to the parameters $\mv \Theta_i^O$ of an output neuron $i\in\mathcal{O}$ is given by
\begin{align}
    &~\nabla_{\scalebox{0.7}{\mv \Theta}_i^O} L_{\mv o^T}(\mv \Theta) = \notag \\
    &~ - \sum_{t=1}^T \mathbb{E}_{p_{\scalebox{0.7}{\mv \Theta}^H}(\scalebox{0.8}{\mv h}_{\mathcal{P}_i,t}|\scalebox{0.8}{\mv h}^{t-1}_{\mathcal{P}_i})}\bigg[\nabla_{\scalebox{0.7}{\mv \Theta}_i^O} \log \big( \text{Tr}(\rho_{i,t}^{\rm I}O_{i,t}) \big) \bigg]. 
\end{align}
This can be estimated by using the local hidden variables $\{\mv h^T_{\mathcal{P}_i}(m)\}_{m=1}^M$ extracted from the corresponding global realizations $\{\mv h^T(m)\}_{m=1}^M$ of the hidden neurons as 
\begin{align}
\hat{\nabla}_{\scalebox{0.7}{\mv \Theta}_i^O} L_{\mv o^T}(\mv \Theta) = - \frac{1}{M} \sum_{m=1}^M \sum_{t=1}^T  \nabla_{\scalebox{0.7}{\mv \Theta}_i^O} \log \big( \text{Tr}(\rho_{i,t}^{\rm I}(m)O_{i,t}(m)) \big), \label{grad5}
\end{align}
where $\rho_{i,t}^{\rm I}(m)$ is the density matrix of the input-output qubit of neuron $i$ during the $m$-th global forward pass, and $O_{i,t}(m)=|\mv o_{i,t}(m)\rangle \langle \mv o_{i,t}(m)|$ is the corresponding $m$-th observable. In this next subsection, we will discuss how to compute the local gradient $\nabla_{\scalebox{0.7}{\mv \Theta}_i^O} \log \big( \text{Tr}(\rho_{i,t}^{\rm I}(m)O_{i,t}(m)) \big)$.
\subsubsection{Hidden neurons}
For any hidden neuron $i\in\mathcal{H}$, using the REINFORCE gradient \cite{8891810}, the gradient with respect to the parameter $\mv \Theta_i^H$ is obtained as
\begin{align}
    \nabla_{\scalebox{0.7}{\mv \Theta}_i^H} L_{\mv o^T}(\mv \Theta) = &~ - \sum_{t=1}^T \mathbb{E}_{p_{\scalebox{0.7}{\mv \Theta}^H}(\scalebox{0.8}{\mv h}_t|\scalebox{0.8}{\mv h}^{t-1})}\bigg[\bigg(\sum_{i\in\mathcal{O}}\log  \text{Tr}(\rho_{i,t}^{\rm I}O_{i,t}) \bigg)  \notag \\
    &~ \times \nabla_{\scalebox{0.7}{\mv \Theta}_i^H} \log \big( \text{Tr}(\rho_{i,t}^{\rm I}H_{i,t}) \big) \bigg], 
\end{align}
where $H_{i,t}=|\mv h_{i,t}\rangle \langle \mv h_{i,t}|$ is the observable corresponding to the realization $\mv h_{i,t}$ of hidden neuron $i\in\mathcal{H}$. Define as 
\begin{align}
    \ell_t(m)=-\sum_{i\in\mathcal{O}} \log  \text{Tr}(\rho_{i,t}^{\rm I}(m)O_{i,t}(m)) \label{log-loss}
\end{align}
the log-loss for the current data point $\mv o^T$ evaluated using the $m$-th realization $\mv h^T(m)$ of the hidden neuron. As seen in Fig.~\ref{local}, the quantities $\{\ell(1), \ldots,\ell(M)\}$, with $\ell(m)=(\ell_1(m), \ldots, \ell_T(m))$, are fed back to all hidden neurons at the end of time $T$, providing global feedback signals to all hidden neurons. Using this information, each neuron $i$ estimates the gradient $\nabla_{\scalebox{0.7}{\mv \Theta}_i^H} L_{\mv o^T}(\mv \Theta)$ as
\begin{align}
\hat{\nabla}_{\scalebox{0.7}{\mv \Theta}_i^H} L_{\mv o^T}(\mv \Theta) = &~ \frac{1}{M} \sum_{m=1}^M \sum_{t=1}^T  \ell_t(m) \times \notag \\ &~\nabla_{\scalebox{0.7}{\mv \Theta}_i^H} \log \big( \text{Tr}(\rho_{i,t}^{\rm I}(m)H_{i,t}(m)) \big). \label{grad6}
\end{align}

\subsection{Local Gradient Estimation}

In this subsection, we focus on estimating the gradient $\nabla_{\scalebox{0.7}{\mv \Theta}_i^O} \log ( \text{Tr}(\rho_{i,t}^{\rm I}(m)O_{i,t}(m)) )$ for any output neuron $i\in \mathcal{O}$ in \eqref{grad5} and the gradient $\nabla_{\scalebox{0.7}{\mv \Theta}_i^H} \log ( \text{Tr}(\rho_{i,t}^{\rm I}(m)H_{i,t}(m)) )$ for any hidden neuron $i\in\mathcal{H}$ in \eqref{grad6}. We write generically such gradients as $\nabla_{\scalebox{0.7}{\mv \Theta}_i} \log \big( \text{Tr}(\rho_{i,t}^{\rm I}S_{i,t}) \big)$ with $S_{i,t}=|\mv s_{i,t}\rangle \langle \mv s_{i,t}|$.
Recall that for each neuron $i\in\mathcal{S}$, the local parameters $\mv \Theta_i=\{\mv w_i,\mv \theta_i\}$ include the synaptic weights $\mv w_i$ and the PQC parameters $\mv \theta_i$. Therefore,  estimating the gradient $\nabla_{\scalebox{0.7}{\mv \Theta}_i} \log \big( \text{Tr}(\rho_{i,t}^{\rm I}S_{i,t}) \big)$ requires evaluating the gradients $\nabla_{\mv w_i} \log \big( \text{Tr}(\rho_{i,t}^{\rm I}S_{i,t}) \big)$ and $\nabla_{\mv \theta_i} \log \big( \text{Tr}(\rho_{i,t}^{\rm I}S_{i,t}) \big)$. These gradients depend on the realization $\mv h^T_{\mathcal{P}_i}$ of the signals of the parents $\mathcal{P}_i$ of node $i$. As described next, we propose to estimate the former using simultaneous perturbation stochastic approximation (SPSA) \cite{spall1998overview}, while the latter is estimated using the parameter shift rule (PSR). The use of the PSR for PQC parameters is standard, while SPSA can be more efficient than individual per-weight perturbations for synaptic weights.

\subsubsection{Gradient with Respect to the Synaptic Weights}
The synaptic weights $\mv w_i=\{\{w_{i,p}^n\}_{p=1}^{P_i}\}_{n=1}^N$ associated with each neuron $i$ pertain to the $NP_i$ links between the $N$ input-output qubits of the $P_i$ pre-synaptic neurons and the corresponding $N$ qubits of neuron $i$. To estimate the gradient $\nabla_{\scalebox{0.7}{\mv w}_{i}} \log \big( \text{Tr}(\rho_{i,t}^{\rm I}S_{i,t}) \big)$ with respect to synaptic weight $\mv w_{i}$, we apply SPSA. 

Accordingly, given the spiking signals of the parent neurons, neuron $i$ runs a number of local forward passes to estimate the expected value $\text{Tr}(\rho_{i,t}^{\rm I}S_{i,t})$ of the observable $S_{i,t}$ obtained with synaptic weights $\mv w_{i} +\epsilon \mv \triangle \mv w_{i,m}$ and $\mv w_i -\epsilon \mv \triangle \mv w_{i,m}$, where $\mv \triangle \mv w_{i,m}$ is the $m$-th random perturbation vector with components drawn independently from the Rademacher distribution, and $\epsilon>0$ represents the magnitude of the perturbation. We generate $M_{\rm syn}$ such independent perturbations $\{\mv \triangle \mv w_{i,m}\}_{m=1}^{M_{\rm syn}}$ for each neuron $i$. Denote as $\rho_{i,t}^{\rm I}(\mv w_i +\epsilon \mv \triangle \mv w_{i,m})$ and $\rho_{i,t}^{\rm I}(\mv w_i -\epsilon \mv \triangle \mv w_{i,m})$ the states of the input-output qubits with perturbed synaptic weights $\mv w_i +\epsilon \mv \triangle \mv w_{i,m}$ and $\mv w_i -\epsilon \mv \triangle \mv w_{i,m}$, respectively. To estimate the corresponding expectations $\text{Tr}(\rho_{i,t}^{\rm I}(\mv w_i +\epsilon \mv \triangle \mv w_{i,m})S_{i,t})$ and $\text{Tr}(\rho_{i,t}^{\rm I}(\mv w_i -\epsilon \mv \triangle \mv w_{i,m})S_{i,t})$, we apply $M_p$ shots. Therefore, the number of local shots for each neuron $i$ for the synaptic weights is $2M_pM_{\rm syn}$. Writing as $\hat{O}^+_{i,m}$ and $\hat{O}^-_{i,m}$ the two estimates obtained with the $m$-th perturbation, the overall local gradient is estimated as
\begin{align}
&~ \hat{\nabla}_{\mv w_{i}} \log \big( \text{Tr}(\rho_{i,t}^{\rm I}S_{i,t}) \big)  =  \frac{1}{M_{\rm syn}} \times \notag   \\ &~ \sum_{m=1}^{M_{\rm syn}} \frac{\log \hat{O}^+_{i,m} - \log \hat{O}^-_{i,m}}{2\epsilon} \mv \triangle \mv w_{i,m}. \label{spsa}
\end{align}

\begin{algorithm}[t!]
\caption{Local learning rule for SQS-based Neural Networks.}\label{alg:sgd}
\SetAlgoNlRelativeSize{-1} 
\SetAlgoNlRelativeSize{0}
\SetEndCharOfAlgoLine{}
\SetKwInput{Input}{\textbf{Input}} 
\SetKwInput{Output}{\textbf{Output}} 

\Input{Training sequence $\mv o^T$ extracted from dataset $\mathcal{D}$}
\Output{Learned model parameters $\mv \Theta$}

\textbf{Initialize} parameters $\mv \Theta$ \;
\underline{\emph{Feedforward sampling:}} \\
Collect $M$ samples of the hidden spiking signals $\mv h^T(1), \ldots, \mv h^T(M)$ and the corresponding global feedback signals $\ell(1), \ldots, \ell(M)$ in \eqref{log-loss}.\;
\underline{\emph{Global feedback:}} \\
Broadcast global learning signals $\ell(1), \ldots, \ell(M)$ to all hidden neurons. \;
\For{each discrete time $t=1,\ldots,T$}{
 \underline{\emph{Local gradient estimation:}} \\
 \For{each neuron $i \in \mathcal{S}$}{
    Compute the gradient $\nabla_{\scalebox{0.7}{\mv \Theta}_i} \log \big( \text{Tr}(\rho_{i,t}^{\rm I}S_{i,t}) \big)$ using \eqref{spsa} or \eqref{proba_theta}. }
}
  \underline{\emph{Parameter update:}} \\
    Compute the gradient for each neuron based on \eqref{grad5} or \eqref{grad6} and perform gradient descent.
\end{algorithm}

\subsubsection{Gradient with Respect to the PQC parameter}
To estimate the gradient $\nabla_{\mv \theta_i} \log \big( \text{Tr}(\rho_{i,t}^{\rm I}S_{i,t}) \big)$ with respect to the PQC parameters $\mv \theta_i$, we apply the PSR to each individual PQC parameter $\theta_{i,j}$. Given the spiking signals from the parent neurons, neuron $i$ performs a series of local forward passes to evaluate the expected value $\text{Tr}(\rho_{i,t}^{\rm I}S_{i,t})$ using perturbed parameters $\theta_{i,j} + \pi/2$ and $\theta_{i,j} -\pi/2$.  For each angle $\theta_{i,j}$, we generate $M_{\rm som}$ independent evaluations using these shifted values. Each expectation is estimated using $M_{\rm p}$ measurement shots per forward pass, leading to a total of $2M_{\rm p}M_{\rm som}$ local shots per angle $\theta_{i,j}$. Let $\hat{O}^+_{i,m}$ and $\hat{O}^-_{i,m}$ denote the measured estimates of the observable $S_{i,t}$ for the $m$-th perturbation with shifted parameters $\theta_{i,j} + \pi/2$ and $\theta_{i,j} -\pi/2$, respectively. Using the chain rule and averaging over the $M_{\rm som}$ repetitions, the gradient is estimated as
\begin{align}
    \nabla_{\theta_{i,j}} \log \big( \text{Tr}(\rho_{i,t}^{\rm I}S_{i,t}) \big) = &~\frac{1}{\text{Tr}(\rho_{i,t}^{\rm I}S_{i,t})} \times \notag \\ &~  \frac{1}{M_{\rm som}} \sum_{m=1}^{M_{\rm som}}\frac{\hat{O}^+_{i,m} - \hat{O}^-_{i,m}}{2}. \label{proba_theta}
\end{align}

The proposed local training procedure is summarized in Algorithm 1. The method performs $M$ global forward passes to obtain spike realizations of hidden neurons and scalar feedback signals, followed by $2TM_{\rm p}|\mathcal{S}|(M_{\rm som}G+M_{\rm syn})$ local passes, with $G$ denoting the number of PQC parameters per neuron and $|\mathcal{S}|$ being the number of neurons. This is contrasted with the $2TMM_{\rm p}|\mathcal{S}|(M_{\rm som}G+M_{\rm syn})$ global forward passes required by conventional zero-th order training.

\section{Experiments} \label{exp}
In this section, we evaluate the performance of the proposed SQSNN model on a variety of classification tasks.

\subsection{Tasks}
We compare the performance of classical and quantum models -- conventional and spiking -- on both standard datasets such as MNIST, FMNIST and KMNIST as in \cite{brand2024quantum}, and on neuromorphic datasets captured via event-driven cameras \cite{gallego2020event}. 

\subsubsection{USPS Binary Classification}
We first consider a simple binary classification task using the USPS handwritten digit dataset, focusing on distinguishing between the digits ``1" and ``7" as in \cite{8891810}.  Each input image is a $16 \times 16$ grayscale representation of a digit and is encoded into the spiking domain using rate encoding \cite{eshraghian2023training}. Accordingly, for each pixel, a spike train of $T=10$ time steps is generated by sampling from a Bernoulli distribution, where the spiking probability is proportional to the pixel's intensity and capped within the range $[0, 0.5]$. The class labels $\{1,7\}$ are also represented using rate encoding during training. Specifically, there are two output neurons, and the target signal $\mv o^T$ for the two output neurons are such that the neuron associated with the correct class emits a spike at each time step \cite{8891810}.

\subsubsection{MNIST, FMNIST, and KMNIST image classification}
Following \cite{brand2024quantum}, we also consider image classification tasks based on the standard MNIST, FMNIST, and KMNIST datasets. Each dataset consists of grayscale images of size $28 \times 28$. Unlike \cite{brand2024quantum}, in order to highlight the capacity of different models to process event-driven data, we encode the inputs for the MNIST, FMNIST and KMNIST datasets using rate encoding, so that the firing probability for the signal encoding each pixel is proportional to its corresponding pixel value. Accordingly, the experiments on these datasets evaluate the models on spike-encoded temporal representations, rather than on the original dense-valued image inputs. The purpose of the experiments on MNIST, FMNIST, and KMNIST is primarily to validate that the proposed SQSNN remains competitive relative to existing spiking and quantum-spiking models under comparable parameter budgets when operating on spike-encoded temporal data. We set $T=5$ discrete time steps.

\subsubsection{Neuromorphic Data}
Finally, we study classification tasks based on a neuromorphic dataset, namely MNIST-DVS \cite{serrano2015poker}, to better showcase the sequence modeling capabilities of SQSNN. The MNIST-DVS dataset consists of labeled spiking signals produced by an event-driven camera with a duration of $T=80$ time steps \cite{chen2023neuromorphic}. Each data point contains $26 \times 26=676$ spike trains, recorded from a dynamic vision sensor (DVS) while it observes handwritten digits moving on a screen. The dataset provides 900 training examples and 100 testing examples per class. For both training and testing as in \cite{wu2021liaf}, only the first $T=20$ time steps are used for classification.

\subsection{Benchmarks}
In the following experiments, we benchmark the proposed SQSNN model against the following alternative models:
\begin{itemize}
    \item \emph{Classical SNN \cite{eshraghian2023training}:} Classical feedforward SNN model based on LIF neurons, trained using surrogate gradient descent with backpropagation.
    \item \emph{QLIF-SNN \cite{brand2024quantum}:} Feedforward neural network implementing QLIF neurons.
    \item \emph{QNN \cite{schuld2020circuit}:} A quantum variational classifier with classical intput and read-out layers. For this model, the input is first encoded using a single classical neural layer, and then processed by a PQC using angle embedding and entangling gates. The outputs are measured and passed to a classical fully connected layer.
\end{itemize}

All models are configured to have approximately the same number of trainable parameters. This is a conventional benchmarking setting in quantum machine learning \cite{abbas2021power, fellner2025quantum} (see Section I). 

\begin{figure}[htp]
\centering
\includegraphics[width=3.1in]{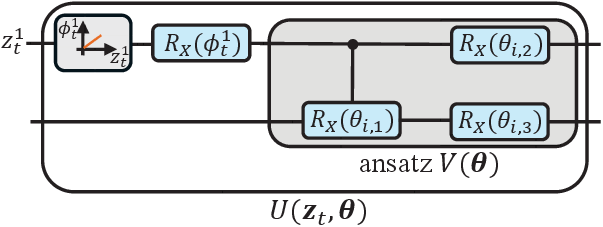}
\caption{Illustration of the quantum circuit for a SQS neuron with $N=1$ and $N_{\rm mem}=1$, where the first line corresponds to input-output qubit, while the second line is for the memory qubit.} \label{circuit}
\end{figure}

\subsection{Architecture}
For all tasks and models, we adopt fully-connected feedforward architectures. The output layers consists of as many neurons as there are classes. For spiking models, we use rate decoding, selecting the class whose corresponding neuron produces the largest number of spikes. In more details, for the USPS dataset, we use fully connected models with a single, output, layer of two neurons. For the other datasets, the models also include a hidden layer. For the SQSNN, there are 1000 hidden neurons for the MNIST, FMNIST and KMNIST datasets, and 256 neurons for the MNIST-DVS dataset, while for the SNN and the QLIF-SNN baselines the number of hidden neurons is adjusted to match the total number of trainable parameters in the SQSNN. The QNN model is also set up to have the same number of parameters. The shallow ansatz in Fig. \ref{circuit} follows the hardware-efficient design principle for near-term devices \cite{cerezo2021variational}.   The learning rate is set to $5 \times 10^{-4}$ for all models.

As illustrated in Fig. \ref{circuit}, for each SQS neuron, we configure the quantum circuit with $N=1$ input-output qubit and $N_{\rm mem}=1$ memory qubit. The choice of $N$ and $N_{\rm mem}$ introduces a trade-off between expressivity and complexity. Increasing these parameters enlarges the Hilbert space explored by the SQS neuron, potentially enabling richer temporal dynamics, but also increasing the likelihood of overfitting \cite{banchi2021generalization, caro2023out}.  Therefore, $N$ and $N_{\rm mem}$ should be treated as architectural hyperparameters and selected through validation procedures that balance accuracy, resource requirements, and generalization performance. In practice, scaling to larger $N$ or $N_{\rm mem}$  would increase the capacity of each SQS neuron, but is currently limited mainly by the exponential cost of classical simulation and by the limited availability of qubits on cloud quantum platforms.

The input currents $\mv z_t$ are encoded into rotation angles $\phi_t^n$ according to the mapping defined in \eqref{recovery}, which determines the input-dependent $R_X(\phi_t^n)$ rotation at each time step. The ansatz $U(\mv z_t, \mv \theta)$ consists of a controlled-RX (CRX) gate followed by single-qubit RX rotations applied independently to each qubit, as in \cite{schuld2020circuit}. The CRX gate is parameterized by a rotation angle $\theta$, and is given by the unitary matrix
\[
\text{CRX}(\theta) =
\begin{bmatrix}
1 & 0 & 0 & 0 \\
0 & 1 & 0 & 0 \\
0 & 0 & \cos(\theta/2) & -i\sin(\theta/2) \\
0 & 0 & -i\sin(\theta/2) & \cos(\theta/2)
\end{bmatrix}
.\]
Both the CRX and RX gates contain trainable parameters that are optimized during learning. Therefore, in this setting, each SQS neuron $i$ has three parameters in vector $\mv \theta_i$ to be learned.

\subsection{Evaluating the Local Learning Rule}

\begin{figure}[htp]
	\centering
	\includegraphics[width=3.1in]{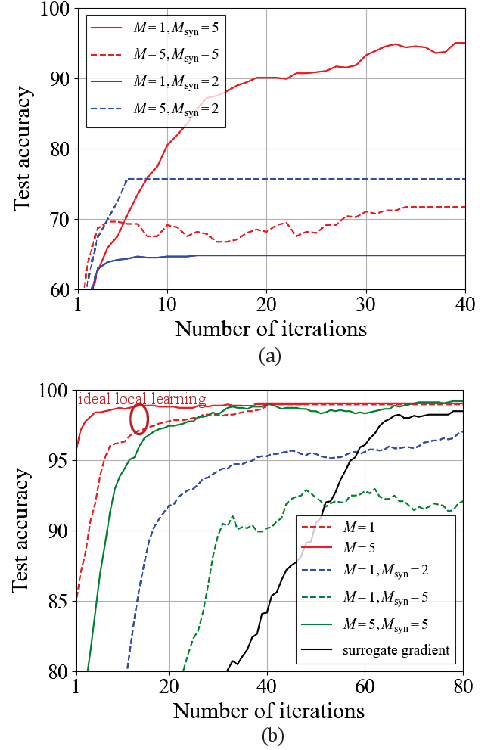}
	\caption{(a) Test accuracy versus training iterations on a NISQ device without QEM: Test accuracy as a function of training iterations for the SQSNN model on the USPS dataset  under the proposed local learning rule implemented on IBM's ibm\_brisbane without QEM.  We vary the number of global passes $M$ and measurement shots for synaptic weights $M_{\rm syn}$, with $M_{\rm som}=1$. (b) Test accuracy versus training iterations on a NISQ device with QEM: Test accuracy as a function of training iterations for the SQSNN model on the USPS dataset  under the proposed local learning rule, implemented on IBM's ibm\_brisbane with zero noise extrapolation algorithm.  The ideal local learning refers to the proposed local learning rule  with a perfect estimation of spiking probability \eqref{prob_spike}, while the surrogate gradient method is an efficient approximate training approach based on classical simulations that is described in the Appendix.}
	\label{result1}
\end{figure} 

We start by evaluating the properties of the proposed local learning rule, focusing on its per-iteration requirements in terms of the number of global forward passes, $M$, and the numbers of local shots at each neuron, namely $2M_{\rm syn}$ for the synaptic weights and $2M_{\rm som}$ for each PQC parameter. 

To this end, we implement the local training rule on the IBM quantum computer ibm\_brisbane using Qiskit \cite{javadi2024quantum}. In Fig. \ref{result1}(a) and Fig. \ref{result1}(b), we report the test accuracy of the SQSNN on the USPS dataset as a function of the training iterations, while varying the parameters $M$ and $M_{\rm syn}$ with $M_{\rm som}=1$ and $M_{\rm p}=5$. Fig. \ref{result1}(a) does not assume quantum error mitigation (QEM), while Fig. \ref{result1}(b) implements QEM using the zero noise extrapolation algorithm provided by the Mitiq library\cite{javadi2024quantum}.

With $M=1$ and $M_{\rm syn}=5$, the SQSNN model is seen to achieve a peak test accuracy of nearly 95\%, even in the absence of QEM. In contrast, configurations with lower shot counts for the synaptic weights, e.g., $M_{\rm syn}=2$, or a higher numbers of global forward passes, e.g., $M=5$, exhibit reduced accuracy. Furthermore, increasing the number of measurement shots from $M_{\text{syn}} = 2$ to $M_{\text{syn}} = 5$ at fixed $M = 5$ leads to a performance drop. We attribute this behavior to the non-stationarity of NISQ hardware. In fact, increasing $M$ or $M_{\text{syn}}$ enlarges the number of circuit executions that are run sequentially to form a single gradient estimate, thereby extending the wall-clock window over which the corresponding measurements are aggregated. Over such windows, the relaxation and dephasing times, as well as the gate and readout error rates of superconducting qubits are known to drift, even within a single calibration cycle \cite{etxezarreta2023multiqubit, wang2023dgr}. Furthermore, the implicit reset inserted between successive circuit executions is imperfect, leaving residual excited-state population that propagates into the following execution as a state-preparation error \cite{event, rost2025long, tan2024extending}.  As a result of these effects, acquiring more samples need not reduce the overall estimation error when this drift is left uncompensated.  Overall, the strong performance of the configuration with $(M=1, M_{\rm syn}=5, M_{\rm som}=1)$ highlights the potential of local learning in hybrid quantum-classical systems, even with current NISQ devices without QEM.

Using the same hardware, Fig. \ref{result1}(b) plots the test accuracy obtained by the SQSNN model on the USPS dataset as a function of training iterations with QEM. We set $M_{\rm som}=1$, and report results for the configurations $(M=1, M_{\rm syn}=2)$,  $(M=1, M_{\rm syn}=5)$ and $(M=5, M_{\rm syn}=5)$.  For reference,  we include the performance of an ideal local training algorithm in which the local gradients \eqref{spsa}-\eqref{proba_theta} are computed exactly. In practice, this would require either a classical model of the neuron or the use of a large number of shots $M_{\rm syn}$ and $M_{\rm som}$. We also evaluate a surrogate gradient-based method inspired by methods for classical SNNs \cite{eshraghian2023training}, in which the spiking probability in \eqref{prob_spike} is replaced with a suitable sigmoid function, enabling the use of backpropagation. This scheme requires a classical simulator, providing a more efficient alternative for implementation on classical processor such as GPUs. We provide a brief description of this approach in Appendix. 

We observe from Fig. \ref{result1}(b) that increasing either the number of global forward passes $M$ or the number of measurement shots $M_{\rm syn}$ leads to improved test accuracy. However, it is generally necessary to increase both numbers in order to obtain the best performance. In particular, with $M=5$ and $M_{\rm syn}=5$, SQSNN obtains a test accuracy around 98\% at around 40 iterations, which matches that of the corresponding ideal local learning rule. Overall, the use of QEM is concluded to further improve the performance of SQSNNs, while also stablizing its performance as a function of the number of global forward passes and shots.

Finally, the surrogate gradient-based strategy---which relies on an efficient classical simulation of the system---provides a close approximation of the performance of the ideal local learning rule at convergence. While the proposed local learning rule is inherently hardware-compatible and intended for on-device adaptation, its direct evaluation requires a quantum processor of sufficient scale to handle benchmarks of practical interest. Therefore, consistent with prior works on quantum spiking neural networks \cite{brand2024quantum}, we henceforth employ classical simulations using a surrogate gradient-based approach to approximate the learning dynamics of the proposed rule (see Appendix). This enables scalable evaluation on larger tasks, while preserving the relevance of the proposed local rule as an efficient learning mechanism for NISQ systems.

\begin{figure}[htp]
	\centering
	\includegraphics[width=3.0in]{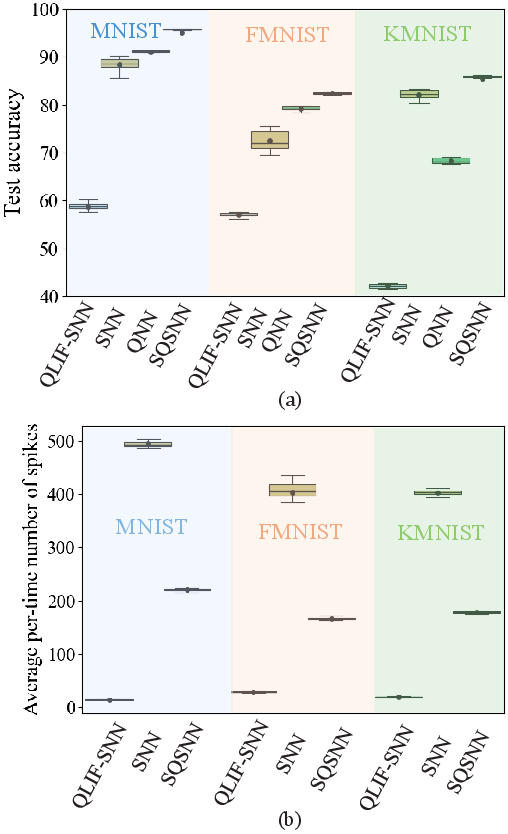}
	\caption{Test accuracy and average number of spike per time of SNN, QNN, QLIF-SNN and SQSNN models on the MNIST, FMNIST, and KMNIST datasets: (a) Box plots of test accuracy achieved by each model on the three datasets. (b) Box plot for the average number of spikes per time step during inference.}
	\label{result2}
\end{figure}

\subsection{Natural Image Tasks}

We now present a  comparative evaluation of the four models, SNN, QNN, QLIF-SNN and SQSNN, on the MNIST, FMNIST, and KMNIST image classification tasks. As shown in Fig.~\ref{result2}(a), the SQSNN model consistently achieves the highest test accuracy across all three datasets, demonstrating its superior capability in extracting relevant features from static input. The SNN also performs well, but lags behind the SQSNN, while the QNN and QLIF-SNN show comparatively lower accuracy than SQSNN, particularly on the more challenging KMNIST dataset.

Following common practice in neuromorphic engineering, we also report the average number of spikes per time step during inference. This quantity represents the aggregate spike count across the entire network, rather than per-neuron spiking activity. We adopt this measure to fairly compare models that differ in neuron count, but are constrained to a similar total parameter budget. As discussed earlier, the number of spikes provides a useful measure of the classical and quantum processing complexity of the SQSNN.  As seen in Fig.~\ref{result2}(b), the QLIF-SNN model exhibits the lowest number of spikes, but at the cost of a substantially lower accuracy. Notably, the SQSNN model achieves the most favorable performance trade-off, maintaining both high accuracy and moderate spiking activity. This balance highlights the potential of the SQSNN for energy-efficient neuromorphic computing, especially in hardware-constrained environment.

\subsection{Neuromorphic Vision Task}
We now evaluate all spiking models, SNN, QLIF-SNN, and SQSNN on a MNIST-DVS dataset. In order to explore the trade-off between accuracy and number of spikes, we add to the training objective \eqref{upper} the regularization term
\begin{align}
     \frac{\lambda}{T}\sum_{t=1}^T \frac{(\sum_i u_{i,t})^2}{\sum_i |u_{i,t}|^2}, \label{regularizer}
\end{align}
where $u_{i,t}$ is a measure related to the spiking rate of neuron $i$ at time $t$, $\lambda>0$ is a hyperparameter, and the sum over index $i$ is evaluated separately for each layer. Specifically, for the SQSNN, the quantity $u_{i,t}$ denotes the spiking probability $p(s_{i,t}=1)$; for the QLIF-SNN, it represents the spiking probability given by multi-shot measurements \cite{brand2024quantum}; and for the classical SNN, it corresponds to the membrane potential \cite{eshraghian2023training}. The regularizer \eqref{regularizer} promotes the sparsity of the vector with entries $\{u_{i,t}\}_i$, encouraging lower spiking rates.

\begin{figure}[htp]
	\centering
	\includegraphics[width=3.1in]{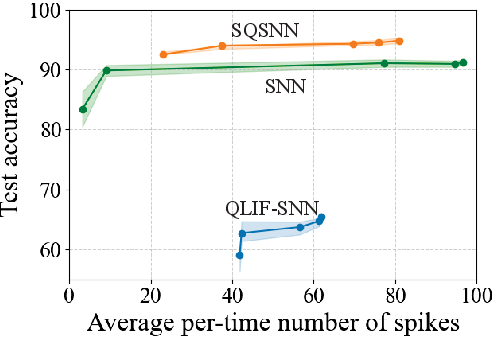}
	\caption{Accuracy versus average number of spikes per time step for SNN, QLIF-SNN and SQSNN on the MNIST-DVS dataset under varying regularization strengths $\lambda$ in \eqref{regularizer}. }
	\label{mnist_dvs}
\end{figure} 

In Fig.~\ref{mnist_dvs}, we compare the test accuracy versus the average number of spikes per time step for the SNN, QLIF-SNN, and SQSNN models on the MNIST-DVS dataset under different regularization strengths $\lambda$. The results demonstrate that the SQSNN consistently outperforms both baseline models, achieving the highest accuracy across a broad range of spiking sparsity levels. Notably, SQSNN maintains high performance even with relatively few spikes, indicating its superior efficiency in processing spatio-temporal event-based data. In contrast, QLIF-SNN exhibits significantly lower accuracy at similar or higher spike counts. This highlights the effectiveness of the quantum-enhanced architecture of SQSNN for extracting discriminative features from neuromorphic inputs.

\subsection{Effect of Multiple Qubits in SQS neurons}

To further examine the role of the multi-qubit SQS neuron design beyond the minimal configuration $N = N_{\rm mem} = 1$, we now revisit the performance on the FMNIST dataset with enlarged quantum dimensions, namely $(N=2, N_{\rm mem}=1)$ and $(N=2, N_{\rm mem}=2)$.

\begin{figure}[htp]
	\centering
\includegraphics[width=3.0in]{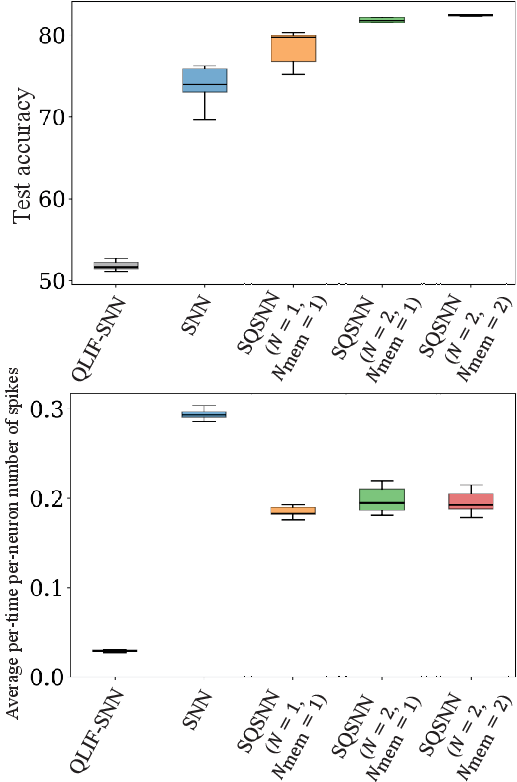}
	\caption{Test accuracy and average per-time per-neuron number of spikes for QLIF-SNN, SNN and SQSNN models with configurations $(N=1, N_{\rm mem}=1)$, $(N=2, N_{\rm mem}=1)$, $(N=2, N_{\rm mem}=2)$ on the rate-encoded FMNIST dataset: (a) Box plots of test accuracy achieved by each model. (b) Box plot for the average per-time number of spikes during inference.} \label{multi}
\end{figure}

Fig.~\ref{multi} shows that enlarging the quantum neuron dimension consistently improves classification accuracy while maintaining sparse spike activity. The minimal SQS configuration $(N=1, N_{\rm mem}=1)$ already outperforms both QLIF-SNN and classical SNN baselines with a lower spike rate. Increasing the number of input qubits to $N=2$ further improves accuracy, and adding an additional memory qubit $(N_{\rm mem}=2)$ achieves the best overall performance. Importantly, these gains are not accompanied by proportional increases in spike density, indicating improved temporal information integration rather than simple activity amplification.

\begin{figure*}[t!]
	\centering
\includegraphics[width=7.0in]{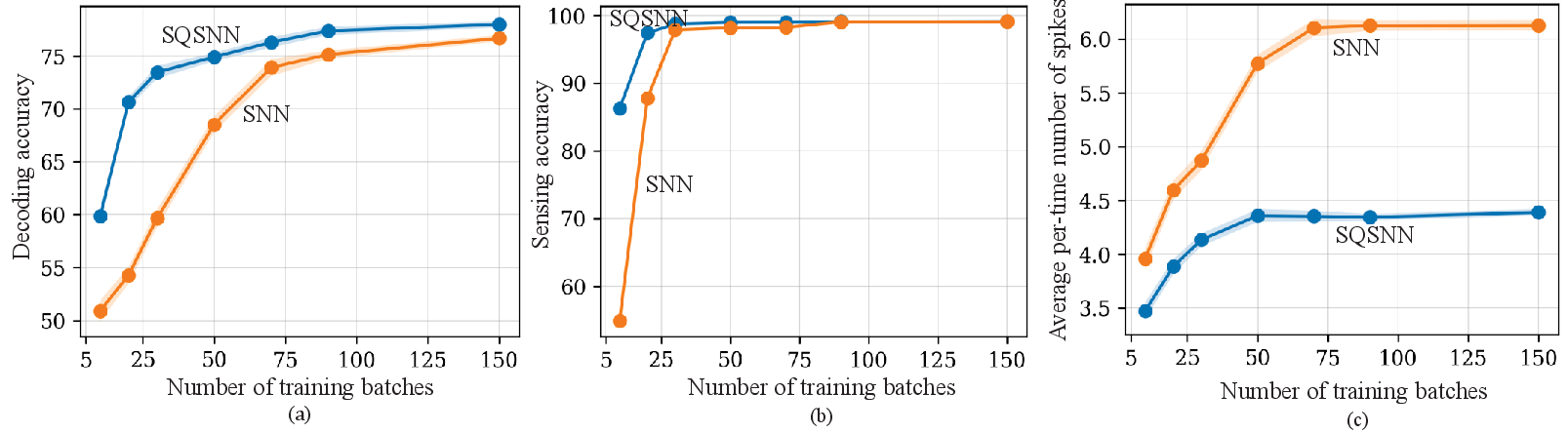}
	\caption{Test decoding accuracy, sensing accuracy, and average number of spikes per time step of the N-ISAC system using SNN and SQSNN, where each training batch contains 128 samples: (a) Decoding accuracy versus the number of training batches. (b) Sensing accuracy versus the number of training batches. (c) Average number of spikes per time step versus the number of training batches.} \label{isac}
\end{figure*}
\subsection{Neuromorphic Integrated Sensing and Communications}

To further demonstrate the practical relevance of the proposed quantum spiking neuron beyond classification benchmarks, we extend the evaluation to a neuromorphic integrated sensing and communications (N-ISAC) scenario \cite{9997098}. 
In this setting, a shared waveform is used simultaneously for digital communication and radar sensing, and the neuromorphic receiver performs joint inference in an event-driven manner.

Specifically, an impulse-radio transmitter employs pulse-position modulation (PPM) to convey a binary sequence $\{x_l\}_{l=1}^{L}$. 
Bit $0$ is represented by a pulse located at the first chip of a slot, whereas bit $1$ is represented by a pulse located at the middle chip within a slot consisting of $2L_b$ chips, where $L_b$ denotes the bandwidth expansion factor. 
The transmitted waveform propagates through a multipath fading channel that may include a radar target with presence indicator $v_l \in \{0,1\}$ at a known delay cell, together with multiple clutter paths. 
The receiver samples the resulting complex baseband signal at the chip rate.

For each slot $l$, the receiver collects the $2L_b$ complex samples into a vector $\mv y_l$ and forms a real-valued representation $\bar{\mv y}_l \in \mathbb{R}^{4L_b}$ by stacking real and imaginary components. 
The temporal sequence $\{\bar{\mv y}_l\}_{l=1}^L$ is then processed by a neuromorphic receiver consisting of spiking neurons with two readout heads. 
These readouts generate a communication decoding decision $\hat{x}_l \in \{0,1\}$ and a sensing decision $\hat{v}_l \in \{0,1\}$ for each slot. 
Training is performed using a weighted sum of cross-entropy losses corresponding to communication and sensing objectives, where a weighting coefficient $\alpha \in [0,1]$ controls the trade-off between the two objectives. In the experiments, we set $\alpha=0.3$. Following the N-ISAC formulation, the decoding accuracy and sensing accuracy are defined as $1/L \sum_{l=1}^L \mathbbm{1}(\hat{x}_l=x_l)$ and $1/L \sum_{l=1}^L \mathbbm{1}(\hat{v}_l=v_l)$ respectively. We refer to \cite{9997098} for further details.

Fig.~\ref{isac} reports the performance obtained when adopting SQSNN and SNN at the receiver, respectively. 
QLIF-SNN is omitted for visual clarity since it consistently achieves the lowest performance among the compared schemes. 
Fig. \ref{isac}(a) shows the decoding accuracy as a function of the number of training batches, where each training batch contains 128 samples. As the amount of training data increases, both receivers exhibit improved decoding performance. However, SQSNN consistently achieves higher decoding accuracy across all training regimes, with the advantage being more pronounced when the number of training batches is small. This suggests that SQSNN exhibits stronger representational efficiency in data-constrained settings.

Fig.~\ref{isac}(b) presents the sensing accuracy under the same training conditions. Sensing performance improves with increasing training batches for both models, while SQSNN maintains a performance advantage in the low training regimes. Fig.~\ref{isac}(c) reports the average number of spikes per time step. While SNN maintains a relatively high spike rate, SQSNN consistently operates with fewer spikes per time step. This demonstrates that the proposed quantum spiking architecture achieves superior sensing and decoding performance with reduced neural activity, reflecting improved event-driven efficiency.

\section{Conclusions} \label{con}
In this work, we have introduced the stochastic quantum spiking (SQS) neuron, a novel quantum computational model that leverages the internal state space of multi-qubit quantum circuits to realize a spiking unit with inherent quantum memory. Unlike prior quantum spiking neuron models that require repeated measurements and classical memory mechanisms, the SQS neuron supports single-shot, event-driven spike generation, making it more aligned with the principles of neuromorphic computing.

We further studied SQS neural networks (SQSNNs) built from SQS neurons, and demonstrated that they can be trained using a local, hardware-efficient learning rule. This rule only requires the application of pertubation-based gradient estimates within each neuron, while steering learning via global feedback signals from the output to the hidden neurons.

By combining the sparse temporal processing efficiency of neuromorphic systems with the rich representational capacity of quantum computing, the proposed SQSNN architecture offers a promising direction for hybrid quantum neuromorphic intelligence. Our results open the door to the development of practical, trainable quantum spiking neural networks that can be implemented on future quantum hardware.

Future work may investigate the implementation on suitable quantum computers such as neutral-atom platforms (see Section \ref{sec:intro}),  the integration with quantum error mitigation, and the implementation on modular multi-core architectures with constraints on the bandwidth between cores as in classical neuromorphic chips  \cite{davies2021advancing}.

\section*{Appendix A: Surrogate Gradient Learning}
This appendix introduces a surrogate gradient-based learning technique enabling the use of backpropagation for a classical simulator of an SQSNN. The approach is inspired by well-established methods for classical SNNs \cite{eshraghian2023training}.

To elaborate, we consider a classification task with $|\mathcal{O}|$ classes, where $\mathcal{O}$ is the set of output neurons. Given an input at each time $t$, the output neurons produce a vector of spikes  $\hat{\mv o}_t=[\hat{\mv o}_{1,t}, \ldots, \hat{\mv o}_{|\mathcal{O}|,t}]$, with $\hat{\mv o}_{i,t}\in\{0,1\}^N$. The spike rate for each output neuron $i\in \mathcal{O}$ over the full sequence of length $T$ is evaluated as $\mv r_i=\frac{1}{T}\sum_{t=1}^T \hat{\mv o}_{i,t}$, providing an estimate of the spiking probability \eqref{upper}. The loss in \eqref{upper} is approximated as the cross-entropy between the spike rates $\mv r_i$ and the target outputs $\mv o_i^T$  
\begin{align}
    \mathcal{L}(\mv \Theta) = &~ \frac{1}{|\mathcal{B}||\mathcal{O}|NT} \sum_{\mv o^T \in \mathcal{B}} \sum_{i=1}^{|\mathcal{O}|} \sum_{n=1}^N \sum_{t=1}^T [-o^n_{i,t} \log(r^n_i) \notag \\
    &~ -(1-o^n_{i,t})\log(1-r^n_i)], \label{losses}
\end{align}
where $\mathcal{B} \subseteq \mathcal{D}$ is a mini-batch.

During the forward process, the output $\mv s_{i,t}$ of each neuron $i$ at each time $t$ is sampled from the Bernoulli distribution with probability $p_{\scalebox{0.7}{\mv \Theta}_i}(\mv s_{i,t}|\mv s_{\mathcal{P}_i\cup \{i\}}^{t-1})$. 
Applying the chain rule, we compute the gradient of the loss \eqref{losses} with respect to model parameters $\mv \Theta_i$ of each neuron $i$ as
\begin{align}
    \frac{\partial \mathcal{L}(\mv \Theta)}{\partial \mv \Theta_i} = \frac{\partial \mathcal{L}(\mv \Theta)}{\partial \mv s_{i,t}} \cdot \frac{\partial \mv s_{i,t}}{\partial p_{\scalebox{0.7}{\mv \Theta}_i}(\mv s_{i,t}|\mv s_{\mathcal{P}_i\cup \{i\}}^{t-1})} \cdot \frac{\partial p_{\scalebox{0.7}{\mv \Theta}_i}(\mv s_{i,t}|\mv s_{\mathcal{P}_i\cup \{i\}}^{t-1})}{\partial \mv \Theta_i}. \label{chain}
\end{align}
However, the spiking signals $\mv s_{i,t}$ are random and thus the chain rule \eqref{chain} cannot be directly used. To address this problem, we replace the $N$-dimensional binary random signal $\mv s_{i,t}$ with a deterministic $N$-dimensional vector taking values in the interval $[0,1]$. Specifically, denoting as $\sigma(\cdot)$ the element-wise sigmoid function $\sigma(x)=1/(1+\exp(-x))$, we replace vector $\mv s_{i,t}$ with the surrogate $\sigma(p_{\scalebox{0.7}{\mv \Theta}_i}(\mv s_{i,t}|\mv s_{\mathcal{P}_i\cup \{i\}}^{t-1}))$, yielding the partial derivative approximation 
\begin{align}
    \frac{\partial \mathcal{L}(\mv \Theta)}{\partial \mv \Theta_i} = &~ \frac{\partial \mathcal{L}(\mv \Theta)}{\partial \mv s_{i,t}} \cdot \frac{\partial \sigma(p_{\scalebox{0.7}{\mv \Theta}_i}(\mv s_{i,t}|\mv s_{\mathcal{P}_i\cup \{i\}}^{t-1}))}{\partial p_{\scalebox{0.7}{\mv \Theta}_i}(\mv s_{i,t}|\mv s_{\mathcal{P}_i\cup \{i\}}^{t-1})} \cdot \notag \\ &~\frac{\partial p_{\scalebox{0.7}{\mv \Theta}_i}(\mv s_{i,t}|\mv s_{\mathcal{P}_i\cup \{i\}}^{t-1})}{\partial \mv \Theta_i}.
\end{align}
This approximation is applied during the backward pass of backpropagation to estimate the gradient $\nabla_{\scalebox{0.7}{\mv \Theta}} \mathcal{L}(\mv \Theta)$.

\small{
\bibliographystyle{ieeetr}
\bibliography{references}
}

\end{document}